\documentclass[lettersize,journal]{IEEEtran}
\usepackage{amsmath,amsfonts}
\usepackage{algorithmic}
\usepackage{algorithm}
\usepackage{array}
\usepackage[caption=false,font=normalsize,labelfont=sf,textfont=sf]{subfig}
\usepackage{textcomp}
\usepackage{stfloats}
\usepackage{url}
\usepackage{verbatim}
\usepackage{graphicx}
\usepackage{cite}
\usepackage{booktabs}
\usepackage{makecell}
\usepackage{gensymb}
\usepackage{hyperref}
\begin{document}

\title{Multi-robot Rigid Formation Navigation via Synchronous Motion\\ and Discrete-time Communication-Control Optimization}

\author{Qun~Yang,
Soung~Chang~Liew$^\ddagger$,~\IEEEmembership{Fellow,~IEEE}
\thanks{Q. Yang, and S. C. Liew are with the Department of Information Engineering, The Chinese University of Hong Kong, Hong Kong SAR, China (e-mail: \{yq020, soung\}@ie.cuhk.edu.hk). }
\thanks{$^\ddagger$S. C. Liew is the corresponding author.}}

\markboth{Journal of \LaTeX\ Class Files,~Vol.~14, No.~8, August~2021}%
{Shell \MakeLowercase{\textit{et al.}}: A Sample Article Using IEEEtran.cls for IEEE Journals}


\maketitle

\begin{abstract}
Rigid-formation navigation of multiple robots is essential for applications such as cooperative transportation. This process involves a team of collaborative robots maintaining a predefined geometric configuration, such as a square, while in motion. For untethered collaborative motion, inter-robot communication must be conducted through a wireless network. Notably, few existing works offer a comprehensive solution for multi-robot formation navigation executable on microprocessor platforms via wireless networks, particularly for formations that must traverse complex curvilinear paths. To address this gap, we introduce a novel "hold-and-hit" communication-control framework designed to work seamlessly with the widely-used Robotic Operating System (ROS) platform. The hold-and-hit framework synchronizes robot movements in a manner robust against wireless network delays and packet loss. It operates over discrete-time communication-control cycles, making it suitable for implementation on contemporary microprocessors. Complementary to hold-and-hit, we propose an intra-cycle optimization approach that enables rigid formations to closely follow desired curvilinear paths, even under the nonholonomic movement constraints inherent to most vehicular robots. The combination of hold-and-hit and intra-cycle optimization ensures precise and reliable navigation even in challenging scenarios. Simulations in a virtual environment demonstrate the superiority of our method in maintaining a four-robot square formation along an S-shaped path, outperforming two existing approaches. Furthermore, real-world experiments validate the effectiveness of our framework: the robots maintained an inter-distance error within $\pm 0.069m$ and an inter-angular orientation error within $\pm 19.15^{\circ}$ while navigating along an S-shaped path at a fixed linear velocity of $0.1 m/s$. Notably, the proposed hold-and-hit framework and optimized nonholonomic motion paradigms are generalizable and extendable to a wide range of multi-robot collaboration problems beyond those studied here. 
\end{abstract}

\begin{IEEEkeywords}
Multi-robot rigid formation navigation, communication-control optimization, cooperative transportation.
\end{IEEEkeywords}

\section{Introduction}\label{sec-I}
\IEEEPARstart{M}{ulti}-Robot Systems (MRS) are the foundation for various applications such as cooperative surveillance, exploration, and transportation \cite{darmanin2017review}. Within this domain, rigid formation control \cite{kanjanawanishkul2011formation} is particularly critical for cooperative transportation tasks. One widely studied approach to rigid formation control is the leader-follower strategy, where follower robots maintain formation by tracking a designated leader’s movements.

In most existing leader-follower schemes, followers adjust their trajectories based solely on their own observations, without direct information from the leader \cite{liu2018survey}. This observation-only approach has a significant drawback: the follower may remain unaware of the leader’s immediate intentions.


\begin{figure}[!t]
\centering
\subfloat[]
{
  \label{subfig1.0}
  \includegraphics[width=0.23\textwidth]{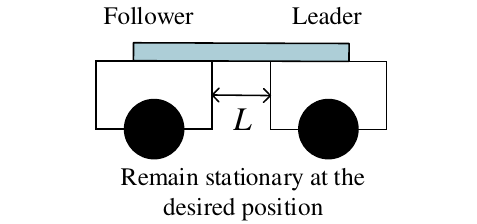}
}
\subfloat[]
{
  \label{subfig1.1}
  \includegraphics[width=0.23\textwidth]{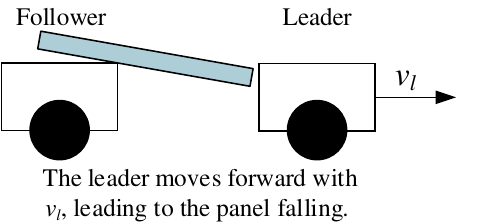}
}
\caption{Cooperative transportation with the leader-follower scheme.}
\label{fig_1}
\end{figure}

Consider the example in Fig. \ref{subfig1.0}. Two robots, a leader and a follower, are initially motionless and holding a panel between them. A distance of $L$ separates the two robots. Now, suppose the leader decides to move forward with a velocity of $v_{l}$. Without prior knowledge of the leader’s intention, the follower remains stationary until the next control cycle, during which it observes a translational deviation of ${v}_{l}T$ from the leader, where $T$ is the cycle duration. If ${v}_{l} T$ is significant, this response lag can disrupt the rigid formation, potentially causing the panel to fall, as illustrated in Fig. \ref{subfig1.1}.

Many MRS prioritize low cost over achieving a fast control loop. For instance, the TurtleBot2, a widely used platform, operates with a recommended control cycle of $T=100ms$\cite{TurtleBot2}. The relatively large $T$, however, makes it challenging for the above approach to achieve rapid synchronized acceleration of the robots within the rigid formation.

Instead of the observation-only approach, the leader can communicate its intentions to the follower before making significant changes in motion. This proactive communication obviates the need for the followers to rely solely on observations to react with an inherent lag in response. Enabling such communication requires the use of a wireless network. An even more direct approach is for the leader to compute the control inputs for the follower and transmit them directly. In this setup, the leader uses an onboard sensor, such as LiDAR, to perceive and measure the relative poses (positions and angular orientations) between itself and the follower. It then combines this information with its own intended motion to calculate the control input for the follower.  This approach departs from the traditional leader-follower paradigm and is referred to in this paper as a \textit{master-slave scheme}. In this scheme, the robot responsible for computing the control inputs acts as the \textit{master}, while the other robots are the \textit{slaves}.

A critical component of the master-slave scheme is wireless communication; however, delays and loss of control inputs due to unstable wireless communication can compromise the integrity of the rigid formation. In the presence of long delays or lost control commands, a follower may fail to update its velocities across multiple consecutive control cycles. This can lead to accumulated pose errors, disrupting the formation. Additionally, in the case of multiple followers, different reception times of the latest control inputs among the followers create further challenges, as they can result in unsynchronized actions across the robots. 

\begin{figure}[!t]
\centering
\includegraphics[width=3.3in]{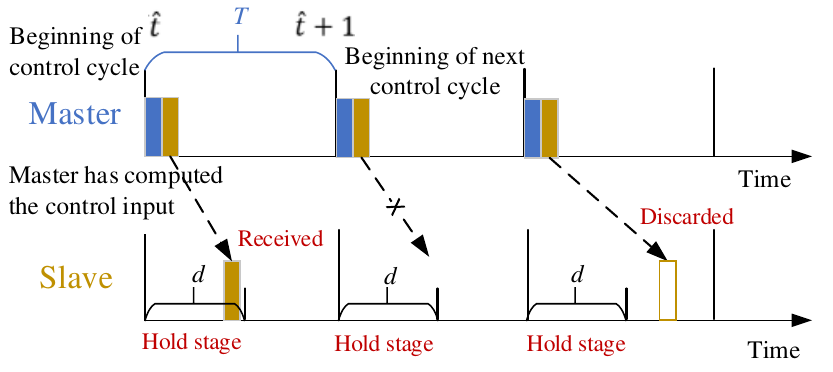}
\caption{Sequence diagram of the hold-and-hit mechanism.}
\label{fig_2}
\end{figure}

To address the wireless loss and synchronization challenges, we put forth a communication-control framework called “\textit{hold-and-hit}”. This framework introduces additional elastic delays to align the delays experienced by all robots. Fig. \ref{fig_2} illustrates the hold-and-hit mechanism. At the start of each cycle, at time instances $\hat{t}=0,1,2...$, the master measures pose errors, computes control inputs, and transmits them to the slaves. A control input is considered successfully conveyed to a slave if it is received before $\hat{t}+d$. If the delay is less than $d$, say $d^{\prime}$, the slave introduces an additional elastic delay of $d-d^{\prime}$ to ensure the total delay equals $d$. In other words, the slave "holds" its action until time $\hat{t}+d$, at which point it applies the newly received control input. This mechanism is named \textit{hold-and-hit}, as the slave holds its action until the designated time and then "hits" by executing the control input. 

\textbf{Addressing Synchronization Challenge:} The objective is for the slaves to update their velocities synchronously at time $\hat{t}+d$, where $d<T$, rather than immediately at $\hat{t}$. The hold-and-hit mechanism ensures all slaves experience the same delay $d$ before applying control inputs, enabling synchronized execution to maintain the integrity of the rigid formation. 

\textbf{Addressing Wireless Loss Challenge: } During the interval $[\hat{t},\hat{t}+d)$, a control input may be lost and retransmitted multiple times by a lower-layer network protocol (e.g., in the Wi-Fi MAC protocol, lost packets can be retransmitted multiple times until they are received). As long as the slave receives a copy before $\hat{t}+d$, the process proceeds smoothly. In particular, the hold mechanism can reduce the packet loss rate by allowing these multiple transmission attempts. If no copy of the control input is received by $\hat{t}+d$, inter-pose errors may accumulate. This is not a fatal problem in our approach, as the master will measure the accumulated error at the next cycle, $\hat{t}+1$, and compute a new control input to correct it, ensuring the system remains on track despite the missed input.

\textbf{Nonholonomic Constraints:} Another significant challenge in rigid-formation nonholonomic mobile robots is navigating along curvilinear paths. Robots with nonholonomic constraints \cite{bloch1989control} lack the freedom to follow arbitrary paths from point A to point B. Severe variations in path curvature can cause rapid accumulation of inter-pose errors, particularly with the nonholonomic constraints limiting the freedom of movement.

Refs \cite{raghuwaiya2018leader} and \cite{alonso2017multi} devised control strategies that take into account the nonholonomic constraints through continuous-time kinematics, where control inputs change continuously over time. However, contemporary robots running on platforms like ROS typically employ discrete-time kinematics, where control inputs change only once per control cycle. Applying continuous-time kinematics to such systems involves approximations, which, in extreme cases, can cause system divergence resulting from error accumulation (as discussed in Section \ref{sec-IV}).

To avoid the continuous-time approximation, this paper puts forth a discrete-time control law, referred to as \textit{Discrete-time Error Minimization} (DEM), which is fully compatible with the hold-and-hit framework and implementable on the ROS platform. To the best of our knowledge, this is the first comprehensive discrete-time framework that ensures rigid formation of nonholonomic robots navigating curvilinear paths via a wireless network. 

Our contributions are threefold:

\begin{enumerate}
    \item \textbf{Hold-and-Hit Framework:} We propose a communication-control strategy called \textit{hold-and-hit} to ensure synchronized motion of collaborative robots, even in the presence of lossy communication.
    \item \textbf{Discrete-Time Control Mechanism (DEM):} We develop a discrete-time control mechanism, \textit{Discrete-time Error Minimization} (DEM), which is compatible with the hold-and-hit framework and accounts for nonholonomic constraints, enabling precise curvilinear path navigation in rigid formations.
    \item \textbf{Validation through Simulations and Experiments:} We validate the effectiveness of hold-and-hit and DEM through extensive simulations and real-world experiments, demonstrating their superiority over existing methods.
\end{enumerate}

\section{Related Work}\label{sec-II}
Table \ref{table1} summarizes and compares prior works on multi-robot formation navigation with our proposed approach, where each column lists a key feature of our work. With reference to the first column, our work stands out from prior studies through its \textit{hold-and-hit} mechanism, which synchronizes the actions and velocity changes of multiple robots as a cohesive group while tolerating network delays and losses. This paper is the first to establish a framework that ensures synchronized motion of collaborative robots despite lossy communication.

\begin{table*}
\centering
\caption{Comparisons of works related to robot formation-navigation problems\label{table1}}
\begin{tabular}{lcccccc}
\toprule
Ref. & \makecell[c]{Synchronized  \\ actions and \\ velocity changes} & \makecell[c]{Discrete-time \\ control model} & \makecell[c]{ Integrated formation \\ and navigation} & \makecell[c]{Nonholonomic \\ Curvilinear-path \\ navigation} & \makecell[c]{Simulation-based \\ investigations} & \makecell[c]{Real-robot \\ experimental \\ investigations} \\
\midrule
our work & \checkmark & \checkmark & \checkmark & \checkmark & \checkmark & \checkmark \\
\cite{cepeda2015formation,yamchi2017formation} & \text{ } & \text{ } & \checkmark & \text{ } & \checkmark & \text{ } \\
\cite{cruz2016leader} & \text{ } & \checkmark & \checkmark & \text{ } & \checkmark & \checkmark \\
\cite{zhao2018time}  & \text{ } & \checkmark & \text{ } & \checkmark & \text{ } & \checkmark \\
\cite{koung2021cooperative} & \text{ } & \text{ } & \text{ } & \checkmark & \checkmark & \checkmark \\
\cite{wang2023optical}  & \text{ } & \text{ } & \checkmark & \text{ } & \checkmark & \checkmark \\
\bottomrule
\end{tabular}
\end{table*}

With reference to the second column, \cite{cepeda2015formation,yamchi2017formation,koung2021cooperative} and \cite{wang2023optical} employed continuous-time controllers that are not fully compatible with contemporary microprocessor platforms, which can only update control inputs at discrete time instances. To overcome this limitation, \cite{cruz2016leader} proposed a discrete-time control approach that approximates nonholonomic leader-follower kinematics, providing a more practical framework for implementation on modern robotic platforms. However, this controller is only effective for navigating robot formations along straight lines and circular paths, and it does not support navigation along arbitrary curvilinear paths (see Section \ref{sec-IV} for further details).

With reference to the third column, many prior works did not adequately address formation and navigation in an integrated manner. For example, \cite{zhao2018time} developed a distributed controller that allowed each robot to navigate its own path without ensuring the collective multi-robot formation during movement. On the other hand, \cite{koung2021cooperative} proposed a hierarchical controller that prioritized formation integrity over navigation. Specifically, when correcting formation errors, navigation would halt, leading to intermittent stop-and-go motions. In contrast, our approach integrates formation control and navigation seamlessly. The robots continue moving along the targeted path while simultaneously correcting any deformation.

The fourth column pertains to nonholonomic curvilinear path navigation. In \cite{zhao2018time}, this was achieved by each individual robot using a distributed controller. In contrast, \cite{koung2021cooperative} enabled curvilinear path navigation by directing the formation’s centroid toward a target, using the centroid’s pose as the control feature. In our approach, the nonholonomic master robot serves as the focal point for navigating along the curvilinear path, while the slave robots are responsible for correcting the inter-pose errors relative to the master.

As for the fifth and sixth columns, all studies except \cite{zhao2018time} employed simulations to evaluate their system performance. All but \cite{cepeda2015formation,yamchi2017formation} realized their proposals on real-robot systems. 

\section{Problem Formulation and Definitions}\label{sec-III}
\begin{figure}
\centering
\includegraphics[width=1.4in]{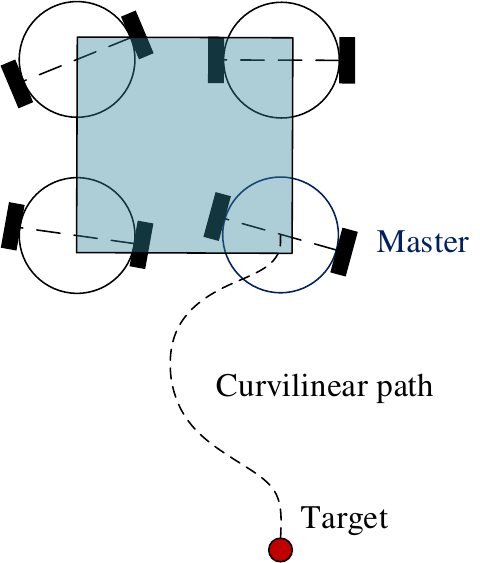}
\caption{Cooperative transportation with four nonholonomic wheeled robots. The blue square represents an object on top of the group. The dashed line is the predefined path towards the target.}
\label{fig_3}
\end{figure}

\begin{figure}
\centering
\subfloat[]
{
  \label{subfig4.0}
  \includegraphics[width=0.2\textwidth]{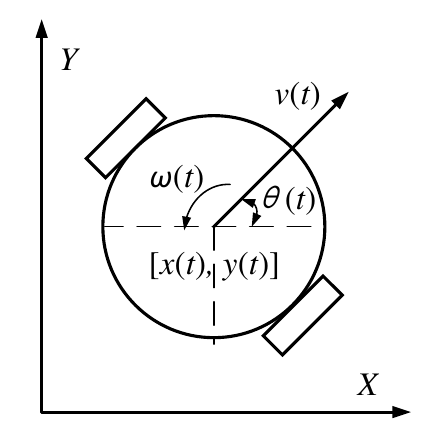}
}
\subfloat[]
{
  \label{subfig4.1}
  \includegraphics[width=0.215\textwidth]{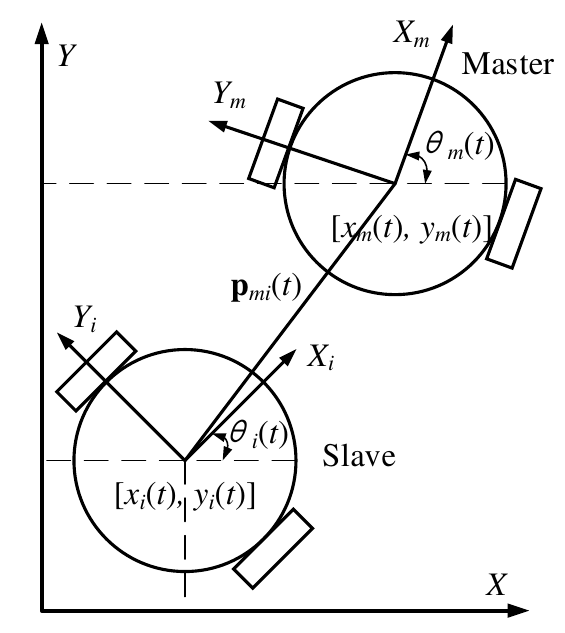}
}
\caption{(a) Differential mobile robot with nonholonomic constraints, (b) The master-slave scheme for rigid formation.}
\label{fig_4}
\end{figure}

Fig. \ref{fig_3} shows a team of four robots tasked with transporting an object from the origin to a target location along a curvilinear path. In our master-slave scheme, the master is responsible for computing the control inputs necessary for the slaves to correct inter-pose errors and maintain the rigid formation.
For simplicity, let us first consider a system with one master and only one slave. With respect to the nonholonomic robot in Fig. \ref{subfig4.0}, we denote the realized pose and velocity vector at time $t\in\mathbb{R}$ by $\mathbf{\textbf{p}}\left(t\right)\in\mathbb{R}^3,\mathbf{\textbf{v}}\left(t\right)\in\mathbb{R}^2$. The pose vector is defined as  $\mathbf{\textbf{p}}\left(t\right)=\left[x\left(t\right),y\left(t\right),\theta\left(t\right) \right]^T$, where  $x\left(t\right),y\left(t\right)$ are the central coordinates (position), and  $\theta\left(t\right)$ is the angular orientation. The velocity vector is defined as $\mathbf{\textbf{v}}\left(t\right)=\left[v\left(t\right),\omega\left(t\right) \right]^T$, where $v\left(t\right),\omega\left(t\right)$ are the linear and angular velocities, respectively. The kinematics of the nonholonomic robot in Fig. \ref{subfig4.0} is
\begin{equation}\label{eq001}
\frac{d\mathbf{\textbf{p}}\left(t\right)}{dt}
= \begin{bmatrix}
\frac{dx\left(t\right)}{dt} \\
\frac{dy\left(t\right)}{dt}  \\
\frac{d\theta\left(t\right)}{dt}
\end{bmatrix}=
\begin{bmatrix}
 \cos{\theta\left(t\right)} & 0 \\
 \sin{\theta\left(t\right)} & 0 \\
 0 & 1
\end{bmatrix}
\begin{bmatrix}
v\left(t\right) \\
\omega\left(t\right)
\end{bmatrix}
\end{equation}

Fig. \ref{subfig4.1} shows a system consisting of a master $m$ and a slave $i$, whose pose and velocity vectors are
\begin{equation}\label{eq002}
\begin{aligned}
\text{ } \text{ } Master:\text{ } \text{ } \mathbf{\textbf{p}}_m\left(t\right)&=\left[x_m\left(t\right),y_m\left(t\right),\theta_m\left(t\right) \right]^T \\
\mathbf{\textbf{v}}_m\left(t\right)&=\left[v_m\left(t\right),\omega_m\left(t\right) \right]^T
\end{aligned}
\end{equation}
\begin{equation}\label{eq003}
\begin{aligned}
Slave: \text{ } \text{ } 
\mathbf{\textbf{p}}_i\left(t\right)&=\left[x_i\left(t\right),y_i\left(t\right),\theta_i\left(t\right) \right]^T \\
\mathbf{\textbf{v}}_i\left(t\right)&=\left[v_i\left(t\right),\omega_i\left(t\right) \right]^T
\end{aligned}
\end{equation}

Let the target velocities corresponding to the control inputs of the robots be $\bar{\mathbf{\textbf{v}}}_m\left(t\right),\bar{\mathbf{\textbf{v}}}_i\left(t\right)\in\mathbb{R}^2$. The realized velocities could deviate from the target velocities due to various sources of errors. In our system, the target velocities only change at discrete time instants $\hat{t}=0,1,2...$. Thus, henceforth, we write the master target velocities as $\bar{\mathbf{\textbf{v}}}_{m,\hat{t}},\hat{t}=0,1,2...$, with the understanding that in continuous time $t$, the target velocities are $\bar{\mathbf{\textbf{v}}}_m\left(t\right)=\bar{\mathbf{\textbf{v}}}_{m,int\left(t\right)}$, where $int\left(t\right)$ is the integer smaller or equal to $t$. For succinctness, we also denote $\bar{\mathbf{\textbf{v}}}_{m,int\left(t\right)}$  by $\bar{\mathbf{\textbf{v}}}_{m,\hat{t}}$ with the understanding that $\hat{t}$ is also used as a shorthand notation for $int\left(t\right)$. 

In an ideal system without noise, we could simply preplan the trajectories of the master and slave to be identical to maintain the rigid formation. That is, the slave changes its velocities at the same instants as the master and $\bar{\mathbf{\textbf{v}}}_{i,\hat{t}}=\bar{\mathbf{\textbf{v}}}_{m,\hat{t}},\hat{t}=0,1,2...$. However, in a system with noise, we cannot do so, as elaborated below. 

For the integrity of rigid formations, we could focus on the errors of the slave with respect to the master. Specifically, we assume the realized velocities of the master are error-free, and as far as the error in rigid formation is concerned, the error is due to the errors in slaves. And it is the slave, not the master, that needs to adjust its velocities to maintain the rigid formation. 

In other words, the velocity errors of the slave are measured against the “standard” velocities of the master. Thus, we write 
\begin{equation}\label{eq004}
\begin{aligned}
&\mathbf{\textbf{v}}_m\left(t\right)=\bar{\mathbf{\textbf{v}}}_{m,\hat{t}} \\
&\mathbf{\textbf{v}}_i\left(t\right)=\bar{\mathbf{\textbf{v}}}_i\left(t\right) + \mathbf{\textbf{n}}_i\left(t\right),
\end{aligned}
\end{equation}
where $\bar{\mathbf{\textbf{v}}}_i\left(t\right)$ is the target velocities of the slave in continuous time, and $\mathbf{\textbf{n}}_i\left(t\right)=\left[n_v\left(t\right),n_\omega\left(t\right)\right]$  is the noise that models the deviations of the slave’s realized velocities from its target velocities. We assume that the system has been calibrated such that $\mathbf{\textbf{n}}_i\left(t\right)$  is a zero-mean symmetric random process. 

The role of $\bar{\mathbf{\textbf{v}}}_i\left(t\right)$ is twofold: (i) to follow the master; (ii) to maintain the rigid formation by compensating for the accumulated errors due to noise. As mentioned earlier, without noise, and if the issue were (i) only, we could simply set $\bar{\mathbf{\textbf{v}}}_i\left(t\right)=\bar{\mathbf{\textbf{v}}}_{m,\hat{t}}$. With noise, (ii) is an issue, and we cannot do so. We next explain how $\bar{\mathbf{\textbf{v}}}_i\left(t\right)$ changes with time in our system. 

Recall that in the hold-and-hit control cycle sequences in Fig. \ref{fig_2}, the master computes the slaves' new velocities at discrete instants $\hat{t}$ and the slaves adopt the new velocities at instants $\hat{t}+d$, $\hat{t}=0,1,2...$. 

Accordingly, in our hold-and-hit system, 
\begin{equation}\label{eq005}
\bar{\mathbf{\textbf{v}}}_i\left(t\right) = \left\{\begin{matrix} 
\bar{\mathbf{\textbf{v}}}_{m,\hat{t}} & for\text{ }t \in [\hat{t},\hat{t}+d)   \\ 
\bar{\mathbf{\textbf{v}}}_{i,\hat{t}} & with \text{ } p \text{ } for \text{ } t \in [\hat{t}+d,\hat{t}+1)   \\
\bar{\mathbf{\textbf{v}}}_{m,\hat{t}} &  with \text{ } q \text{ } for \text{ } t \in [\hat{t}+d,\hat{t}+1)
\end{matrix}\right.
\end{equation}

The idea of the above equation is as follows. For the time interval $[\hat{t},\hat{t}+d)$, the slave adopts the preplanned trajectory as the master by following its velocities. The time interval $[\hat{t}+d,\hat{t}+1)$ is for the slave to perform error correction. The above equation captures the fact that the wireless network may not be $100\%$ reliable. With probability $p$, a slave receives the instruction from the master, in which case it sets its target velocities as instructed by the master, $\bar{\mathbf{\textbf{v}}}_{i,\hat{t}}$. With probability $q=1-p$, the slave does not receive the instruction from the master and therefore will continue to follow the pre-planned velocities $\bar{\mathbf{\textbf{v}}}_{m,\hat{t}}$ in the time interval $[\hat{t}+d,\hat{t}+1)$. We emphasize that the error-correction velocities $\bar{\mathbf{\textbf{v}}}_{i,\hat{t}}$, if received, are adopted only for the time interval $[\hat{t}+d,\hat{t}+1)$. 

We next describe how the master computes the error-correction velocities $\bar{\mathbf{\textbf{v}}}_{i,\hat{t}}$ under the robot’s nonholonomic constraint. The computation of $\bar{\mathbf{\textbf{v}}}_{i,\hat{t}}$ is based on the measured formation error between the master and the slave. 

With reference to Fig. \ref{subfig4.1}, let the relative pose between the master and the slave in terms of the slave's frame of reference be $\mathbf{\textbf{p}}_{mi}\left(t\right)=\left[{x}_{mi}\left(t\right),{y}_{mi}\left(t\right),{\theta}_{mi}\left(t\right)\right]^T$. We can write
\begin{equation}\label{eq007}
\mathbf{\textbf{p}}_{mi}\left(t\right)=
\begin{bmatrix}
 \cos{\theta}_{i}\left(t\right) & \sin{\theta}_{i}\left(t\right) & 0 \\
 -\sin{\theta}_{i}\left(t\right) & \cos{\theta}_{i}\left(t\right) & 0 \\
 0 & 0 & 1
\end{bmatrix}
\begin{bmatrix}
{x}_{m}\left(t\right)-{x}_{i}\left(t\right) \\
{y}_{m}\left(t\right)-{y}_{i}\left(t\right) \\
{\theta}_{m}\left(t\right)-{\theta}_{i}\left(t\right)
\end{bmatrix}
\end{equation}

Let $\mathbf{\textbf{p}}^d_{mi}=\left[{x}^d_{mi},{y}^d_{mi},{\theta}^d_{mi}\right]^T$ be the desired relative pose. For a rigid formation, $\mathbf{\textbf{p}}^d_{mi}$ is a constant vector when expressed in terms of the slave’s frame of reference. The relative pose error is 
\begin{equation}\label{eq008}
\mathbf{\textbf{e}}\left(t\right)=
\begin{bmatrix}
{x}\left(t\right) \\
{y}\left(t\right) \\
{\theta}\left(t\right)
\end{bmatrix}=\mathbf{\textbf{p}}_{mi}\left(t\right)-\mathbf{\textbf{p}}^d_{mi}
\end{equation}

Eqn. (\ref{eq008}) is merely an expression for the error $\mathbf{\textbf{e}}\left(t\right)$; it does not imply that $\mathbf{\textbf{e}}\left(t\right)$ can be observed at all time by the master. During the time interval $[\hat{t},\hat{t}+1)$, the master only observes the relative pose error $\mathbf{\textbf{e}}_{\hat{t}}$ at instant $\hat{t}$. Thus, $\mathbf{\textbf{e}}_{\hat{t}}$ is not a random variable because the master has observed the relative pose error at instant $\hat{t}$ , eliminating any uncertainty about the error. However, $\mathbf{\textbf{e}}\left(t\right)$ for $t\in(\hat{t},\hat{t}+1)$  (i.e., between the observations at $\hat{t}$ and $\hat{t}+1$) is a random variable. 

To maintain the rigid formation while navigating a team of robots, the master must compute $\bar{\mathbf{\textbf{v}}}_{i,\hat{t}}$ based on $\mathbf{\textbf{e}}_{\hat{t}}$ to correct for the projected error at instant $\hat{t}+1$. Thus, the design of $\bar{\mathbf{\textbf{v}}}_{i,\hat{t}}$ should consider the accumulated error during the interval $[\hat{t},\hat{t}+1)$. 

In this paper, we choose to minimize a weighted 2-norm of expected error. Specifically, the optimization problem of interest to us is
\begin{equation}\label{eq009}
\begin{aligned}
    & \min_{\mathbf{\bar{\mathbf{\textbf{v}}}}_{i,\hat{t}}<\mathbf{\bar{\mathbf{\textbf{v}}}}_{max}}\left\|\mathbf{\textbf{W}}E\left[\mathbf{\textbf{e}}\left(t\right)\right]\right \|_2  \\
    & \text{ } \text{ } for \text{ } t=1\_,2\_,...
\end{aligned}
\end{equation}
where $\left\|\text{ }\right \|_2$ is the 2-norm of a vector, $E\left[\text{ }\right]$ is the expectation of a random variable, $\mathbf{\bar{\mathbf{\textbf{v}}}}_{max}$ is a maximum bound imposed on the target velocity $\mathbf{\bar{\mathbf{\textbf{v}}}}_{i,\hat{t}}$, $\mathbf{\textbf{W}}=diag\left(\sqrt{w_x},\sqrt{w_y},\sqrt{w_\theta}\right)$ is a weight matrix,  and $t=1\_,2\_,...$ are the instants just before the next master’s observations at $\hat{t}=1,2,...$. For example, for interval $[2,3)$, we want to minimize the weighted 2-norm expected error at instant 3 based on the observed error at instant 2, $\mathbf{\textbf{e}}_2$; and for interval $[3,4)$, we want to minimize the weighted 2-norm expected error at instant 4 based on the observed error at instant 3.

To compute the optimal $\mathbf{\bar{\mathbf{\textbf{v}}}}_{i,\hat{t}}$ in (\ref{eq009}), we next delve into the evolution of $E\left[\mathbf{\textbf{e}}\left(t\right)\right]$ in one control cycle of our hold-and-hit mechanism. The error $\mathbf{\textbf{e}}\left(t\right)$ is a function of $\bar{\mathbf{\textbf{v}}}_i\left(t\right)$ in (5). To be explicit, we write $\mathbf{\textbf{e}}\left(t\right)$ as $\mathbf{\textbf{e}}\left(\bar{\mathbf{\textbf{v}}}_i\left(t\right),t\right)$. The formation error $\mathbf{\textbf{e}}\left(t\right)$ for $t\in[\hat{t}+d,\hat{t}+1)$, is
\begin{equation}\label{eq010}
\begin{aligned}
&\mathbf{\textbf{e}}\left(t\right) = \mathbf{\textbf{e}}_{\hat{t}} + \int_{\hat{t}}^{\hat{t}+d}
\frac{d\mathbf{\textbf{e}}\left(\bar{\mathbf{\textbf{v}}}_i\left(t\right),\tau\right)}{d\tau} d\tau + 
\int_{\hat{t}+d}^{t}\frac{d\mathbf{\textbf{e}}\left(\bar{\mathbf{\textbf{v}}}_i\left(t\right),\tau\right)}{d\tau} d\tau  \\
&= \left\{\begin{matrix} 
\mathbf{\textbf{e}}_{\hat{t}} + \int_{\hat{t}}^{\hat{t}+d}
\frac{d\mathbf{\textbf{e}}(\bar{\mathbf{\textbf{v}}}_{m,\hat{i}},\tau)}{d\tau} d\tau + 
\int_{\hat{t}+d}^{t}\frac{d\mathbf{\textbf{e}}(\bar{\mathbf{\textbf{v}}}_{i,\hat{t}},\tau)}{d\tau} d\tau & with \text{ } p   \\ 
\mathbf{\textbf{e}}_{\hat{t}} + \int_{\hat{t}}^{t}\frac{d\mathbf{\textbf{e}}(\bar{\mathbf{\textbf{v}}}_{m,\hat{i}},\tau)}{d\tau} d\tau  & with \text{ } q   
\end{matrix}\right.
\end{aligned}
\end{equation}

Taking the expected value of (\ref{eq010}) results in (\ref{eq011}) on the top of the next page.
\begin{figure*}[h]
\begin{equation}\label{eq011}
\begin{aligned}
E\left[\mathbf{\textbf{e}}\left(t\right)\right]&=\mathbf{\textbf{e}}_{\hat{t}}+\int_{\hat{t}}^{\hat{t}+d}\frac{dE\left[\mathbf{\textbf{e}}(\bar{\mathbf{\textbf{v}}}_{m,\hat{t}},\tau)\right]}{d\tau} d\tau + p\int_{\hat{t}+d}^{t}\frac{dE\left[\mathbf{\textbf{e}}(\bar{\mathbf{\textbf{v}}}_{i,\hat{t}},\tau)\right]}{d\tau} d\tau + q\int_{\hat{t}+d}^{t}\frac{dE\left[\mathbf{\textbf{e}}(\bar{\mathbf{\textbf{v}}}_{m,\hat{t}},\tau)\right]}{d\tau} d\tau   \\
&=\underbrace{\mathbf{\textbf{e}}_{\hat{t}}+\overbrace{\int_{\hat{t}}^{t}\frac{dE\left[\mathbf{\textbf{e}}(\bar{\mathbf{\textbf{v}}}_{m,\hat{t}},\tau)\right]}{d\tau} d\tau}^{a} - \overbrace{p\int_{\hat{t}+d}^{t}\frac{dE\left[\mathbf{\textbf{e}}(\bar{\mathbf{\textbf{v}}}_{m,\hat{t}},\tau)\right]}{d\tau} d\tau}^{b}}_{\alpha} + \underbrace{p\int_{\hat{t}+d}^{t}\frac{dE\left[\mathbf{\textbf{e}}(\bar{\mathbf{\textbf{v}}}_{i,\hat{t}},\tau)\right]}{d\tau} d\tau}_{\beta}
\end{aligned}
\end{equation}
\end{figure*}

In (\ref{eq011}), $p$ is the probability that the slave receives $\bar{\mathbf{\textbf{v}}}_{i,\hat{t}}$ from the master before instant $\hat{t}+d$. Given the preplanned $\bar{\mathbf{\textbf{v}}}_{m,\hat{t}}$, $E\left[\mathbf{\textbf{e}}(\bar{\mathbf{\textbf{v}}}_{m,\hat{t}},\tau)\right]$ is a constant vector, and therefore the term $\alpha$ is a constant vector with no unknowns. Specifically, the optimizing $\bar{\mathbf{\textbf{v}}}_{i,\hat{t}}$ in (\ref{eq009}) is contained in term $\beta$ in ($\ref{eq011}$) only. 

Let us focus on the term $\beta$. Differentiating $\mathbf{\textbf{e}}\left(t\right)$ in (\ref{eq008}) gives
\begin{equation}\label{eq012}
\begin{aligned}
\frac{d\mathbf{\textbf{e}}\left(t\right)}{dt} = & \frac{d\mathbf{\textbf{p}}_{mi}\left(t\right)}{dt} =  \begin{bmatrix}
\frac{d{x}_{mi}\left(t\right)}{dt} \\
\frac{d{y}_{mi}\left(t\right)}{dt} \\
\frac{d{\theta}_{mi}\left(t\right)}{dt}
\end{bmatrix} = \\
\begin{bmatrix}
 0 & \omega_{i}\left(t\right) & 0 \\
 -\omega_{i}\left(t\right) & 0 & 0 \\
 0 & 0 & 0
\end{bmatrix}
& \begin{bmatrix}
{x}_{mi}\left(t\right) \\
{y}_{mi}\left(t\right) \\
{\theta}_{mi}\left(t\right)
\end{bmatrix}   + \begin{bmatrix}
v_m\left(t\right)\cos{{\theta}_{mi}\left(t\right)}-v_i\left(t\right)    \\
v_m\left(t\right)\sin{{\theta}_{mi}\left(t\right)}   \\
\omega_{m}\left(t\right) - \omega_{i}\left(t\right) 
\end{bmatrix}
\end{aligned}
\end{equation}

Let $\left[x_d,y_d,\theta_d\right]^T$ be the relative pose error at instant $\hat{t}+d$. Applying (\ref{eq004}) to the expression of $\omega_{m}\left(t\right) - \omega_{i}\left(t\right)$ in (\ref{eq012}) gives the evolution of $\theta_{mi}\left(t\right)$ for $t\in[\hat{t}+d,\hat{t}+1)$:
\begin{equation}\label{eq013}
\begin{aligned}
&\frac{d{\theta}_{mi}\left(t\right)}{dt} = \bar{\omega}_{m,\hat{t}} - \bar{\omega}_{i,\hat{t}} - n_\omega\left(t\right)\\
\theta_{mi}\left(t\right) &= \theta_d + \left(\bar{\omega}_{m,\hat{t}} - \bar{\omega}_{i,\hat{t}} \right)\left(t-\hat{t}-d\right)-N_\omega\left(t\right),
\end{aligned}
\end{equation}
where $N_\omega\left(t\right)=\int_{\hat{t}+d}^{t}n_\omega\left(t\right)dt$. Let $\mathbf{\textbf{N}}_i\left(t\right)=\left[N_v\left(t\right),N_\omega\left(t\right)\right]=\int_{\hat{t}+d}^{t}\mathbf{\textbf{n}}_i\left(t\right)dt$ be the incremental translational and angular velocity errors accumulated over the interval $[\hat{t}+d,t)$. Given the assumption of $\mathbf{\textbf{n}}_i\left(t\right)$ being a zero-mean symmetric random process, $\mathbf{\textbf{N}}_i\left(t\right)$ is therefore also a zero-mean symmetric random process. Applying (\ref{eq004}) and (\ref{eq013}) to the first two rows of (\ref{eq012}), we have
\begin{equation}\label{eq014}
\begin{aligned}
\frac{d{x}_{mi}\left(t\right)}{dt} &= \left[\bar{\omega}_{i,\hat{t}}+n_\omega\left(t\right)\right]y_{mi}\left(t\right)  - \bar{v}_{i,\hat{t}} - n_v\left(t\right) + \\
&  \bar{v}_{m,\hat{t}}\cos{\left[\theta_d+(\bar{\omega}_{m,\hat{t}}-\bar{\omega}_{i,\hat{t}})(t-\hat{t}-d)-N_\omega\left(t\right)\right]} \\
\frac{d{y}_{mi}\left(t\right)}{dt}  &= \left[-\bar{\omega}_{i,\hat{t}}-n_\omega\left(t\right)\right]x_{mi}\left(t\right) + \\
&  \bar{v}_{m,\hat{t}}\sin{\left[\theta_d+(\bar{\omega}_{m,\hat{t}}-\bar{\omega}_{i,\hat{t}})(t-\hat{t}-d)-N_\omega\left(t\right)\right]}
\end{aligned}
\end{equation}

Note that although $y_{mi}\left(t\right)$ in (\ref{eq014}) may depend on the noise $n_\omega\left(t\right)$ prior to time $t$, the noise at time $t$ does not have an immediate effect on the position $y_{mi}\left(t\right)$ yet (it only has an effect on $\frac{dy_{mi}\left(t\right)}{dt}$)\footnote{We assume the noise is attributed to the imperfection in the robot and the relative difference calibration in the master and slave robots, and the position has no bearing on $n_\omega\left(t\right)$.}. 

We can obtain the following equation for $\frac{dE\left[\mathbf{\textbf{e}}\left(t\right)\right]}{dt}=\frac{dE\left[\mathbf{\textbf{p}}_{mi}\left(t\right)\right]}{dt}$ by taking the expected value of (\ref{eq014}) and applying the zero-mean symmetric assumptions of $\mathbf{\textbf{n}}_i\left(t\right)$ and $\mathbf{\textbf{N}}_i\left(t\right)$ (in the following, we make use of the fact that $E\left[\sin{N_\omega\left(t\right)}\right]=0$ because $\mathbf{\textbf{N}}_i\left(t\right)$ is zero-mean symmetric and $\sin{()}$ is an odd function):
\begin{equation}\label{eq015}
\begin{aligned}
& \text{ } \text{ } \text{ } \text{ } \text{ } \text{ } \text{ } \text{ } \text{ } \text{ } \text{ } 
\begin{bmatrix}
\frac{dE\left[x_{mi}\left(t\right)\right]}{dt} \\
\frac{dE\left[y_{mi}\left(t\right)\right]}{dt}  \\
\frac{dE\left[\theta_{mi}\left(t\right)\right]}{dt}
\end{bmatrix}=
\begin{bmatrix}
\bar{\omega}_{i,\hat{t}}E\left[y_{mi}\left(t\right)\right] -\bar{v}_{i,\hat{t}}  \\
 -\bar{\omega}_{i,\hat{t}}E\left[x_{mi}\left(t\right)\right]  \\
 \bar{\omega}_{m,\hat{t}}-\bar{\omega}_{i,\hat{t}} 
\end{bmatrix} +  \\
& \begin{bmatrix}
\bar{v}_{m,\hat{t}}\cos{\left[\theta_d+(\bar{\omega}_{m,\hat{t}}-\bar{\omega}_{i,\hat{t}})(t-\hat{t}-d)\right]}E\left[\cos{N_\omega\left(t\right)}\right]  \\
\bar{v}_{m,\hat{t}}\sin{\left[\theta_d+(\bar{\omega}_{m,\hat{t}}-\bar{\omega}_{i,\hat{t}})(t-\hat{t}-d)\right]}E\left[\cos{N_\omega\left(t\right)}\right]   \\
0
\end{bmatrix}
\end{aligned}
\end{equation}

If we further assume that $N_\omega\left(t\right)$ in (\ref{eq015}) is Gaussian with variance $\sigma_\omega^2\left(t\right)$, then $E\left[\cos{N_\omega\left(t\right)}\right]=e^{-\frac{\sigma_\omega^2\left(t\right)}{2}}$. Thus, we have

\begin{equation}\label{eq016}
\begin{aligned}
& \text{ } \text{ } \text{ } \begin{bmatrix}
\frac{dE\left[x_{mi}\left(t\right)\right]}{dt} \\
\frac{dE\left[y_{mi}\left(t\right)\right]}{dt}  \\
\frac{dE\left[\theta_{mi}\left(t\right)\right]}{dt}
\end{bmatrix}=\begin{bmatrix}
 0 & \bar{\omega}_{i,\hat{t}} & 0 \\
 -\bar{\omega}_{i,\hat{t}} & 0 & 0 \\
 0 & 0 & 0
\end{bmatrix}
\begin{bmatrix}
E\left[x_{mi}\left(t\right)\right] \\
E\left[y_{mi}\left(t\right)\right] \\
E\left[\theta_{mi}\left(t\right)\right]
\end{bmatrix} + \\
& \begin{bmatrix}
\bar{v}_{m,\hat{t}}\cos{\left[\theta_d+(\bar{\omega}_{m,\hat{t}}-\bar{\omega}_{i,\hat{t}})(t-\hat{t}-d)\right]}e^{-\frac{\sigma_\omega^2\left(t\right)}{2}}-\bar{v}_{i,\hat{t}}  \\
  \bar{v}_{m,\hat{t}}\sin{\left[\theta_d+(\bar{\omega}_{m,\hat{t}}-\bar{\omega}_{i,\hat{t}})(t-\hat{t}-d)\right]}e^{-\frac{\sigma_\omega^2\left(t\right)}{2}}  \\
 \bar{\omega}_{m,\hat{t}}-\bar{\omega}_{i,\hat{t}} 
\end{bmatrix}
\end{aligned}
\end{equation}

Typically, the accumulation of noise power in (\ref{eq016}) is linear with respect to time, implying that $\sigma_\omega^2\left(t\right)=\rho t$, where $\rho\in\mathbb{R}^+$. Note that $\rho$ depends on the deviation between the slave's realized angular velocity and its target angular velocity, and it can be measured statistically. We express (\ref{eq016}) in a more compact form over the interval $[\hat{t}+d,t)$ as
\begin{equation}\label{eq017}
\frac{dE\left[\mathbf{\textbf{p}}_{mi}\left(t\right)\right]}{dt}=\mathbf{\textbf{A}}_iE\left[\mathbf{\textbf{p}}_{mi}\left(t\right)\right]+\mathbf{\textbf{b}}_{mi}\left(t\right),
\end{equation}
where 
\begin{equation}\label{eq018}
\begin{aligned}
& \text{ } \text{ } \text{ } \text{ } \text{ } \text{ } \text{ } \text{ } \text{ } \text{ } \text{ } \text{ } \text{ } \mathbf{\textbf{A}}_i = \begin{bmatrix}
 0 & \bar{\omega}_{i,\hat{t}} & 0 \\
 -\bar{\omega}_{i,\hat{t}} & 0 & 0 \\
 0 & 0 & 0
\end{bmatrix}, 
\mathbf{\textbf{b}}_{mi}\left(t\right) = \\
& \begin{bmatrix}
\bar{v}_{m,\hat{t}}\cos{\left[\theta_d+(\bar{\omega}_{m,\hat{t}}-\bar{\omega}_{i,\hat{t}})(t-\hat{t}-d)\right]}e^{-\frac{\rho t}{2}}-\bar{v}_{i,\hat{t}}  \\
  \bar{v}_{m,\hat{t}}\sin{\left[\theta_d+(\bar{\omega}_{m,\hat{t}}-\bar{\omega}_{i,\hat{t}})(t-\hat{t}-d)\right]}e^{-\frac{\rho t}{2}}  \\
 \bar{\omega}_{m,\hat{t}}-\bar{\omega}_{i,\hat{t}} 
\end{bmatrix}
\end{aligned}
\end{equation}

Deriving the theoretical expression of $E\left[\mathbf{\textbf{p}}_{mi}\left(t\right)\right]$ with respect to $\left(\bar{v}_{m,\hat{t}},\bar{\omega}_{m,\hat{t}},\bar{v}_{i,\hat{t}},\bar{\omega}_{i,\hat{t}},t\right)$ from (\ref{eq017}) is challenging due to its nonlinearity and non-homogeneity. A general approach to determine $E\left[\mathbf{\textbf{p}}_{mi}\left(t\right)\right]$ from (\ref{eq017}) involves finding a particular solution $\mathbf{\textbf{p}}_{\beta}^*\left(t\right),t\in[\hat{t}+d,\hat{t}+1)$ of $E\left[\mathbf{\textbf{p}}_{mi}\left(t\right)\right]$. We use the method of \textit{Variation of Parameters} \cite{lakshmikantham2019method} to derive the particular solution $\mathbf{\textbf{p}}_{\beta}^*\left(t\right)$ of $E\left[\mathbf{\textbf{p}}_{mi}\left(t\right)\right]$ in (\ref{eq017}), based on the particular solution of its corresponding homogeneous system. We define the corresponding homogeneous system of (\ref{eq017}) over the interval $[\hat{t}+d,t)$ as
\begin{equation}\label{eq019}
\frac{dE\left[\mathbf{\textbf{p}}_{mi}\left(t\right)\right]}{dt}=\mathbf{\textbf{A}}_iE\left[\mathbf{\textbf{p}}_{mi}\left(t\right)\right]
\end{equation}

Let $\mathbf{\textbf{p}}^*_{(t,1)},\mathbf{\textbf{p}}^*_{(t,2)},\mathbf{\textbf{p}}^*_{(t,3)}$ be the particular solutions of (\ref{eq019}) corresponding to three eigenvalues of $\mathbf{\textbf{A}}_i$. We obtain these solutions through the eigenvector approach and Euler’s formula \cite{moya2011differential}. Let $\mathbf{\textbf{P}}\left(t\right)=\left[\mathbf{\textbf{p}}^*_{(t,1)}\text{ }\mathbf{\textbf{p}}^*_{(t,2)}\text{ }\mathbf{\textbf{p}}^*_{(t,3)}\right]$, we obtain
\begin{equation}\label{eq020}
\mathbf{\textbf{P}}\left(t\right)=\begin{bmatrix}
 0 & \sin{(\bar{\omega}_{m,\hat{t}}t)} & \cos{(\bar{\omega}_{m,\hat{t}}t)} \\
 0 & \cos{(\bar{\omega}_{m,\hat{t}}t)} & -\sin{(\bar{\omega}_{m,\hat{t}}t)} \\
 1 & 0 & 0
\end{bmatrix}
\end{equation}

Let the particular solution $\mathbf{\textbf{p}}_{\beta}^*\left(t\right)$ of $E\left[\mathbf{\textbf{p}}_{mi}\left(t\right)\right]$ in (\ref{eq017}) take the following form \cite{lakshmikantham2019method}:
\begin{equation}\label{eq021}
\mathbf{\textbf{p}}_{\beta}^*\left(t\right)=\mathbf{\textbf{P}}\left(t\right)\mathbf{\textbf{v}}\left(t\right)=\left[\mathbf{\textbf{p}}^*_{(t,1)}\text{ }\mathbf{\textbf{p}}^*_{(t,2)}\text{ }\mathbf{\textbf{p}}^*_{(t,3)}\right]\begin{bmatrix}
v_1  \\
v_2  \\
v_3
\end{bmatrix},
\end{equation}
where $\mathbf{\textbf{v}}\left(t\right)$ is a transforming vector. The problem of finding $\mathbf{\textbf{p}}_{\beta}^*\left(t\right)$ is transformed into solving $\mathbf{\textbf{v}}\left(t\right)$ in (\ref{eq021}). Applying (\ref{eq021}) to (\ref{eq017}), we have
\begin{equation}\label{eq022}
\frac{d\mathbf{\textbf{v}}\left(t\right)}{dt}=\mathbf{\textbf{P}}\left(t\right)^{-1}\mathbf{\textbf{b}}_{mi}\left(t\right)\
\end{equation}

Eqn. (\ref{eq022}) holds because the matrix $\mathbf{\textbf{P}}\left(t\right)$ in (\ref{eq020}) is reversible. Integrating (\ref{eq022}) with respect to $t$ for $t\in[\hat{t}+d,\hat{t}+1)$, yields
\begin{equation}\label{eq023}
\mathbf{\textbf{v}}\left(t\right)=\int_{\hat{t}+d}^{t}\mathbf{\textbf{P}}\left(\tau\right)^{-1}\mathbf{\textbf{b}}_{mi}\left(\tau\right)d\tau
\end{equation}

Applying $\mathbf{\textbf{P}}\left(t\right)$ in (\ref{eq020}) and $\mathbf{\textbf{v}}\left(t\right)$ in (\ref{eq023}) to (\ref{eq021}), and defining $t^\prime=t-(\hat{t}+d)$ , where $t^\prime \in [0,1-d)$, we have 
\begin{equation}\label{eq024}
\begin{aligned}
& \text{ } \text{ } \text{ } \text{ } \text{ }  \text{ } \text{ } \mathbf{\textbf{p}}_{\beta}^*\left(t\right) =\begin{bmatrix}
-\frac{\bar{v}_{i,\hat{t}}}{\bar{\omega}_{i,\hat{t}}}\sin{\bar{\omega}_{i,\hat{t}}}t^\prime    \\
\frac{\bar{v}_{i,\hat{t}}}{\bar{\omega}_{i,\hat{t}}} - \frac{\bar{v}_{i,\hat{t}}}{\bar{\omega}_{i,\hat{t}}}\cos{\bar{\omega}_{i,\hat{t}}}t^\prime      \\
\left(\bar{\omega}_{m,\hat{t}}-\bar{\omega}_{i,\hat{t}}\right)t^\prime
\end{bmatrix}   \\ + 
& \begin{bmatrix}
\frac{2\rho\bar{v}_{m,\hat{t}}}{\rho^2 + 4\bar{\omega}_{m,\hat{t}}^2}\cos{\psi}_{d,i,t^\prime} - \frac{4\bar{v}_{m,\hat{t}}\bar{\omega}_{m,\hat{t}}}{\rho^2 + 4\bar{\omega}_{m,\hat{t}}^2}\sin{\psi}_{d,i,t^\prime}   \\
\frac{4\bar{v}_{m,\hat{t}}\bar{\omega}_{m,\hat{t}}}{\rho^2 + 4\bar{\omega}_{m,\hat{t}}^2}\cos{\psi}_{d,i,t^\prime} + \frac{2\rho\bar{v}_{m,\hat{t}}}{\rho^2 + 4\bar{\omega}_{m,\hat{t}}^2}\sin{\psi}_{d,i,t^\prime}     \\
0
\end{bmatrix}  \\ + 
&  \begin{bmatrix}
- \frac{2\rho\bar{v}_{m,\hat{t}}e^{-\frac{\rho t^\prime}{2}}}{\rho^2 + 4\bar{\omega}_{m,\hat{t}}^2}\cos{\gamma} + \frac{4\bar{v}_{m,\hat{t}}\bar{\omega}_{m,\hat{t}}e^{-\frac{\rho t^\prime}{2}}}{\rho^2 + 4\bar{\omega}_{m,\hat{t}}^2}\sin{\gamma}        \\
- \frac{4\bar{v}_{m,\hat{t}}\bar{\omega}_{m,\hat{t}}e^{-\frac{\rho t^\prime}{2}}}{\rho^2 + 4\bar{\omega}_{m,\hat{t}}^2}\cos{\gamma} - \frac{2\rho\bar{v}_{m,\hat{t}}e^{-\frac{\rho t^\prime}{2}}}{\rho^2 + 4\bar{\omega}_{m,\hat{t}}^2}\sin{\gamma}      \\
0
\end{bmatrix}
\end{aligned}
\end{equation}
where ${\psi}_{d,i,t^\prime}=\theta_d - \bar{\omega}_{i,\hat{t}}t^\prime$, and $\gamma=\theta_d + \left(\bar{\omega}_{m,\hat{t}} - \bar{\omega}_{i,\hat{t}}\right)t^\prime$. In (\ref{eq024}), $\mathbf{\textbf{p}}_{\beta}^*\left(t\right)$ characterizes the evolution of $E\left[\mathbf{\textbf{e}}\left(t\right)\right]$ with respect to $\left(\bar{v}_{m,\hat{t}},\bar{\omega}_{m,\hat{t}},\bar{v}_{i,\hat{t}},\bar{\omega}_{i,\hat{t}},t\right)$ for $t\in[\hat{t}+d,\hat{t}+1)$. Furthermore, $p\mathbf{\textbf{p}}_{\beta}^*\left(t\right)$ is equivalent to the term $\beta$ in (\ref{eq011}). We emphasize that (\ref{eq024}) characterizes the variation of the formation error with respect to the applied target velocities during an intra-cycle interval. We thus can use (\ref{eq024}) to derive the expression for the term $\alpha$ in (\ref{eq011}). By substituting $\bar{\mathbf{\textbf{v}}}_{i,\hat{t}} = \bar{\mathbf{\textbf{v}}}_{m,\hat{t}}$ into (\ref{eq024}), we have the expression for sub-term $b$ in the term $\alpha$ of (\ref{eq011}), which is given by (\ref{eq025}).
\begin{equation}\label{eq025}
\begin{aligned}
& \int_{\hat{t}+d}^{t}\frac{dE\left[\mathbf{\textbf{e}}(\bar{\mathbf{\textbf{v}}}_{m,\hat{t}},\tau)\right]}{d\tau} d\tau = \mathbf{\textbf{p}}_{\alpha,b}^*\left(t\right) = 
 \begin{bmatrix}
-\frac{\bar{v}_{m,\hat{t}}}{\bar{\omega}_{m,\hat{t}}}\sin{\bar{\omega}_{m,\hat{t}}}t^\prime      \\
 - \frac{\bar{v}_{m,\hat{t}}}{\bar{\omega}_{m,\hat{t}}}\cos{\bar{\omega}_{m,\hat{t}}}t^\prime       \\
0
\end{bmatrix} \\ 
& + \begin{bmatrix}
\frac{2\rho\bar{v}_{m,\hat{t}}}{\rho^2 + 4\bar{\omega}_{m,\hat{t}}^2}\cos{\psi}_{d,m,t^\prime} - \frac{4\bar{v}_{m,\hat{t}}\bar{\omega}_{m,\hat{t}}}{\rho^2 + 4\bar{\omega}_{m,\hat{t}}^2}\sin{\psi}_{d,m,t^\prime}  \\
\frac{\bar{v}_{m,\hat{t}}}{\bar{\omega}_{m,\hat{t}}} + \frac{4\bar{v}_{m,\hat{t}}\bar{\omega}_{m,\hat{t}}}{\rho^2 + 4\bar{\omega}_{m,\hat{t}}^2}\cos{\psi}_{d,m,t^\prime} + \frac{2\rho\bar{v}_{m,\hat{t}}}{\rho^2 + 4\bar{\omega}_{m,\hat{t}}^2}\sin{\psi}_{d,m,t^\prime}   \\
0
\end{bmatrix} \\
& + \begin{bmatrix}
- \frac{2\rho\bar{v}_{m,\hat{t}}e^{-\frac{\rho t^\prime}{2}}}{\rho^2 + 4\bar{\omega}_{m,\hat{t}}^2}\cos{\theta_d} + \frac{4\bar{v}_{m,\hat{t}}\bar{\omega}_{m,\hat{t}}e^{-\frac{\rho t^\prime}{2}}}{\rho^2 + 4\bar{\omega}_{m,\hat{t}}^2}\sin{\theta_d}   \\
 - \frac{4\bar{v}_{m,\hat{t}}\bar{\omega}_{m,\hat{t}}e^{-\frac{\rho t^\prime}{2}}}{\rho^2 + 4\bar{\omega}_{m,\hat{t}}^2}\cos{\theta_d} - \frac{2\rho\bar{v}_{m,\hat{t}}e^{-\frac{\rho t^\prime}{2}}}{\rho^2 + 4\bar{\omega}_{m,\hat{t}}^2}\sin{\theta_d}    \\
 0
\end{bmatrix}
\end{aligned}
\end{equation}

In (\ref{eq025}), ${\psi}_{d,m,t^\prime}=\theta_d - \bar{\omega}_{m,\hat{t}}t^\prime$. Recall that the master observes $\mathbf{\textbf{e}}_{\hat{t}} = \left[x_i,y_i,\theta_i\right]^T$ at instant $\hat{t}$, and $\theta_{mi}\left(t\right)$ evolves according to (\ref{eq013}). Thus, the relative angular orientation error $\theta_d$ at instant $\hat{t} + d$ can be estimated by
\begin{equation}\label{eq026}
\begin{aligned}
\theta_d &= \theta_{\hat{t}} + E\left[(\bar{\omega}_{m,\hat{t}} - \bar{\omega}_{i,\hat{t}})d - N_\omega\left(t\right)\right] \\
& = \theta_{\hat{t}} + (\bar{\omega}_{m,\hat{t}} - \bar{\omega}_{i,\hat{t}})d
\end{aligned}
\end{equation}

Applying $\bar{\mathbf{\textbf{v}}}_{i,\hat{t}} = \bar{\mathbf{\textbf{v}}}_{m,\hat{t}}$ to (\ref{eq026}) and substituting (\ref{eq026}) into ${\psi}_{d,m,t^\prime}$, we obtain $\theta_d=\theta_i$ and ${\psi}_{d,m,t^\prime} = \theta_{\hat{t}} - \bar{\omega}_{m,\hat{t}}t^\prime = {\psi}_{\hat{t},m,t^\prime}$. Therefore, the RHS of (\ref{eq025}), which is equivalent to the sub-term $b$ in term $\alpha$ of (\ref{eq011}), is a constant vector with no unknowns. To derive the expression for sub-term $a$ in term $\alpha$ of (\ref{eq011}), we redefine $t\in[\hat{t},\hat{t} + 1)$ in (\ref{eq024}), and apply $\theta_d = \theta_i$ and $\bar{\mathbf{\textbf{v}}}_{i,\hat{t}} = \bar{\mathbf{\textbf{v}}}_{m,\hat{t}}$ to (\ref{eq024}), yielding
\begin{equation}\label{eq027}
\begin{aligned}
& \int_{\hat{t}}^{t}\frac{dE\left[\mathbf{\textbf{e}}(\bar{\mathbf{\textbf{v}}}_{m,\hat{t}},\tau)\right]}{d\tau} d\tau = \mathbf{\textbf{p}}_{\alpha,a}^*\left(t\right) = \begin{bmatrix}
-\frac{\bar{v}_{m,\hat{t}}}{\bar{\omega}_{m,\hat{t}}}\sin{\bar{\omega}_{m,\hat{t}}}(t^\prime + d)   \\
- \frac{\bar{v}_{m,\hat{t}}}{\bar{\omega}_{m,\hat{t}}}\cos{\bar{\omega}_{m,\hat{t}}}(t^\prime + d)    \\
0
\end{bmatrix}   \\
& + \begin{bmatrix}
\frac{2\rho\bar{v}_{m,\hat{t}}}{\rho^2 + 4\bar{\omega}_{m,\hat{t}}^2}\cos{\psi}_{\hat{t},m,t^\prime + d} - \frac{4\bar{v}_{m,\hat{t}}\bar{\omega}_{m,\hat{t}}}{\rho^2 + 4\bar{\omega}_{m,\hat{t}}^2}\sin{\psi}_{\hat{t},m,t^\prime + d}  \\
\frac{\bar{v}_{m,\hat{t}}}{\bar{\omega}_{m,\hat{t}}} + \frac{4\bar{v}_{m,\hat{t}}\bar{\omega}_{m,\hat{t}}}{\rho^2 + 4\bar{\omega}_{m,\hat{t}}^2}\cos{\psi}_{\hat{t},m,t^\prime +d} + \frac{2\rho\bar{v}_{m,\hat{t}}}{\rho^2 + 4\bar{\omega}_{m,\hat{t}}^2}\sin{\psi}_{\hat{t},m,t^\prime+d}   \\
0
\end{bmatrix}   \\
& + \begin{bmatrix}
- \frac{2\rho\bar{v}_{m,\hat{t}}e^{-\frac{\rho (t^\prime+d)}{2}}}{\rho^2 + 4\bar{\omega}_{m,\hat{t}}^2}\cos{\theta_{\hat{t}}} + \frac{4\bar{v}_{m,\hat{t}}\bar{\omega}_{m,\hat{t}}e^{-\frac{\rho (t^\prime + d)}{2}}}{\rho^2 + 4\bar{\omega}_{m,\hat{t}}^2}\sin{\theta_{\hat{t}}}  \\
- \frac{4\bar{v}_{m,\hat{t}}\bar{\omega}_{m,\hat{t}}e^{-\frac{\rho (t^\prime+d)}{2}}}{\rho^2 + 4\bar{\omega}_{m,\hat{t}}^2}\cos{\theta_{\hat{t}}} - \frac{2\rho\bar{v}_{m,\hat{t}}e^{-\frac{\rho (t^\prime+d)}{2}}}{\rho^2 + 4\bar{\omega}_{m,\hat{t}}^2}\sin{\theta_{\hat{t}}}    \\
0
\end{bmatrix},
\end{aligned}
\end{equation}
where ${\psi}_{\hat{t},m,t^\prime+d}=\theta_{\hat{t}} - \bar{\omega}_{m,\hat{t}}(t^\prime + d)$. The RHS of (\ref{eq027}) is also a constant vector with no unknowns, similar to that of (\ref{eq025}). Applying (\ref{eq024}), (\ref{eq025}) and (\ref{eq027}) to (\ref{eq009}), the $\bar{\mathbf{\textbf{v}}}_{i,\hat{t}}$ that minimizes the projected error at instant $\hat{t} + 1$ is given by the following optimization problem:
\begin{equation}\label{eq028}
\begin{aligned}
    &arg\min_{\mathbf{\bar{\mathbf{\textbf{v}}}}_{i,\hat{t}}<\mathbf{\bar{\mathbf{\textbf{v}}}}_{max}}  \\
    & \left\|\left(\mathbf{\textbf{W}}\left[\mathbf{\textbf{e}}_{\hat{t}} + \mathbf{\textbf{p}}_{\alpha,a}^*(\hat{t} + 1) - p\mathbf{\textbf{p}}_{\alpha,b}^*(\hat{t} + 1) + p\mathbf{\textbf{p}}_{\beta}^*(\hat{t} + 1)  \right]\right)^T\right \|_2  \\
    & for \text{ } t=1\_,2\_,....
\end{aligned}
\end{equation}

In (\ref{eq028}), the intra-cycle minimization defines a discrete-time control law, referred to as \textit{Discrete-time Error Minimization} (DEM). Furthermore, it is straightforward to extend (\ref{eq028}) to the MRS depicted in Fig. \ref{fig_3}, which consists of one master and three slaves, by setting $i=1,2,3$.

For our simulation and real experiment studies, we implemented the hold-and-hit mechanism and the DEM control law on ROS. We refer readers to Appendix A for implementation details.

\section{Simulations}\label{sec-IV}
This section evaluates the hold-and-hit framework and the DEM control law in a virtual environment, focusing on two criteria: the ability to maintain rigid formation during motion and robustness under lossy wireless communication. We first compare the effectiveness of our approach with two existing methods in maintaining rigid formation under error-free communication conditions. After that, we assess the robustness of our approach in non-error-free wireless networks.

\subsection{Comparative Evaluation}\label{subsec-IV(A)}
We compare our approach with two existing methods: the discrete-time leader-follower method in \cite{cruz2016leader} and the continuous-time leader-follower method in \cite{wang2023optical}. All three methods compute robot control inputs based on their relative poses. Since the two benchmark methods did not account for communication loss in their control design, we only conducted simulations under error-free communication for this comparative study.

Under error-free communication, $p=1$ in (\ref{eq028}). We examine how the formation error varies with respect to $\bar{\mathbf{\textbf{v}}}_{m,\hat{t}}$ and $T$. Increasing $\bar{\mathbf{\textbf{v}}}_{m,\hat{t}}$ or $T$ makes it more challenging to maintain rigid formation during motion. As will be shown shortly, our approach outperforms the benchmark methods for various $\bar{\mathbf{\textbf{v}}}_{m,\hat{t}}$ and $T$.

\begin{figure}[!t]
\centering
\includegraphics[width=3.3in]{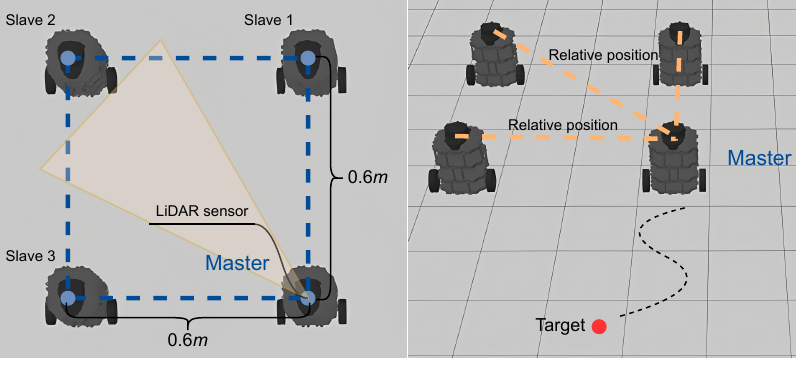}
\caption{Four TurtleBots were arranged in a square formation. Using a LiDAR sensor, the master measured the relative positions between itself and the slaves, and computed the control inputs $\bar{\mathbf{\textbf{v}}}_{i,\hat{t}}$ for the slaves.}
\label{fig_5}
\end{figure}

\begin{figure}[!t]
\centering
\includegraphics[width=1.8in]{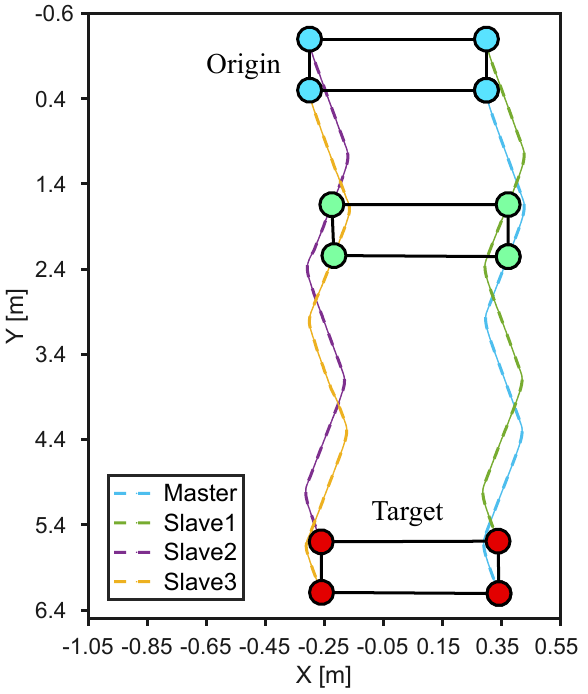}
\caption{The blue dashed line is the S-shaped path for the master to navigate.}
\label{fig_5a}
\end{figure}

We simulated a team of four TurtleBots \cite{TurtleBot3} in a virtual environment using Gazebo. As illustrated in Fig. \ref{fig_5}, the robots were initially arranged in a square formation with length $0.6m$ on each side. The robots were tasked with navigating along the S-shaped path shown in Fig. \ref{fig_5a} as a rigid formation, moving from an origin to a target location. The robots need to maintain their relative positions and angular orientations during navigation. 

As shown in Fig. \ref{fig_5}, the robot at the lower-right corner was the master, while the other robots, indexed 1, 2, and 3, were the slaves. Each robot was unaware of its absolute position in the virtual environment. The master used a LiDAR sensor to detect the relative positions of the slaves, while each robot was equipped with odometry to measure its current angular orientation. The master calculated control inputs for each slave based on the perceived formation errors and transmitted error-correction velocities to the slaves. Simultaneously, the master executed a sequence of preplanned target velocity vectors $\left[\bar{\mathbf{\textbf{v}}}_{m,\hat{t}}\right]$, $\hat{t}=0,1,2...$, to navigate along the S-shaped path. The maximum target velocity for the slaves was set to $\bar{\mathbf{\textbf{v}}}_{max}=1.5max\left|\bar{\mathbf{\textbf{v}}}_{m,\hat{t}}\right|$ -- a limit is imposed to prevent abrupt changes in motion for smooth operation. It is worth noting that the methods in \cite{cruz2016leader} and \cite{wang2023optical} are fully compatible with this setup, with the master acting as the leader and the slaves as followers. However, the next paragraph explains a caveat.

In hold-and-hit, the execution of the computed control inputs $\bar{\mathbf{\textbf{v}}}_{i,\hat{t}}$ is delayed by a constant $d$ to ensure synchronized executions by all robots under the possibility of varying communication delays between the master and different slaves. The methods in 
in \cite{cruz2016leader} and \cite{wang2023optical}, however, assume the control inputs computed at instant $\hat{t}$ are also executed at instant $\hat{t}$ -- i.e., they assume the communication between the master and slaves is perfect and the delays are zero, an ideal but not practical setting. In contrast, we assume a more realistic setting with non-zero communication delays. To cater to different communication delays among the slaves, hold-and-hit synchronizes the slaves' executions of control inputs to occur at $\hat{t}+d$. In other words, in this benchmark study, we give the two existing methods an “unfair” advantage over our system by assuming instantaneous communication in their systems. We will provide results demonstrating that our system has superior performance despite its more realistic non-zero delay setting. 

We investigated several combinations of preplanned target velocity vectors $\bar{\mathbf{\textbf{v}}}_{m,\hat{t}}$ and control cycle durations $T$. These combinations were divided into two groups to separately assess the impact of $\bar{\mathbf{\textbf{v}}}_{m,\hat{t}}$ and $T$ on formation errors.

\textbf{Group 1} examines how formation error varies with the target velocity vector $\bar{\mathbf{\textbf{v}}}_{m,\hat{t}}=\left[\bar{v}_{m,\hat{t}},\bar{\omega}_{m,\hat{t}}\right]^T$, with $T$ fixed to $0.1s$ for all experiments. For each experiment, we fixed the linear velocity $\bar{v}_{m,\hat{t}}$ throughout the navigation. However, for different experiments, we adopted different fixed $\bar{v}_{m,\hat{t}}$. We investigated three $\bar{v}_{m,\hat{t}}$ in three different experiments: $\bar{v}_{m,\hat{t}}=0.05,0.1$, and $0.2m/s$. 

\begin{figure}[!t]
\centering
\includegraphics[width=2.5in]{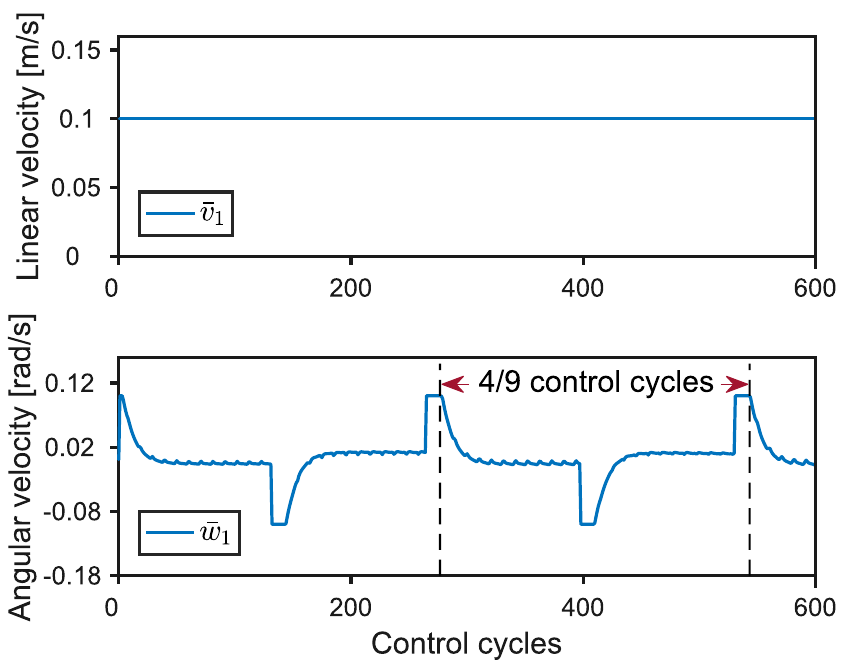}
\caption{The setting of $\bar{v}_{m,\hat{t}}=0.1m/s$ and $\bar{\omega}_{m,\hat{t}}\in\left[-0.1,0.1\right]rad/s$ for a particular experiment.}
\label{fig_6}
\end{figure}

For each experiment, we need to vary the angular velocity $\bar{\omega}_{m,\hat{t}}$ to navigate the team of robots along the S-shaped path. For experiments with $\bar{v}_{m,\hat{t}}=0.1m/s$, $\bar{\omega}_{m,\hat{t}}$ varies within the range $\left[-0.1,0.1\right]rad/s$ in order to navigate along the S-shaped path. Fig. \ref{fig_6} details the setting of $\bar{v}_{m,\hat{t}}$ and $\bar{\omega}_{m,\hat{t}}$. For experiments with $\bar{v}_{m,\hat{t}}=0.05m/s$, $\bar{\omega}_{m,\hat{t}}$ varies within the range $\left[-0.05,0.05\right]rad/s$; for $\bar{v}_{m,\hat{t}}=0.1m/s$, $\bar{\omega}_{m,\hat{t}}\in\left[-0.1,0.1\right]rad/s$; and for $\bar{v}_{m,\hat{t}}=0.2m/s$, $\bar{\omega}_{m,\hat{t}}\in\left[-0.2,0.2\right]rad/s$. Note from Fig. \ref{fig_6} that $\bar{\omega}_{m,\hat{t}}$ varies in a periodical manner with a period equal to $4/9$ of the total number of control cycles. {\hfill $\square$}

\textbf{Group 2} examines how the control cycle duration $T$ affects formation error, with the linear velocity $\bar{v}_{m,\hat{t}}$ fixed to $0.1m/s$ for all experiments. For each experiment, we fixed $T$ throughout. However, for different experiments, different fixed $T$ were adopted. We investigated three $T$ in three different experiments: $T=0.05,0.1$, and $0.2s$. For each experiment, we varied the angular velocity $\bar{\omega}_{m,\hat{t}}$ periodically within the range $\left[-0.1,0.1\right]rad/s$, following the pattern shown in the example in Fig. \ref{fig_6}. {\hfill $\square$}

Note that the product of the total number of control cycles, the target linear velocity $\bar{v}_{m,\hat{t}}$, and the control cycle duration $T$ is a constant for all experiments. This is because the same S-shaped path is being traversed by the robots in all experiments, and the product is the total distance travelled. 

Other parameters in the DEM control law include the delay $d$, which we set to $0.5$ to strike a balance between the reliability of control input delivery and the smooth correction of formation errors -- larger $d$ allows more retransmission attempts at the WiFi MAC layer to achieve higher reliability, while smaller $d$ allows more time to correct for a pose error to ensure smoother operation. We set the weight matrix $\mathbf{\textbf{W}}$ in (\ref{eq028}) to $\mathbf{\textbf{W}}=diag(1,1,1)$, indicating that the relative angular orientation error is considered as important as the relative position error in rigid formation navigation. We measured the noise power coefficient $\rho=1.4153\times10^{-5}$ based on $400$ single-robot driving experiments conducted in the same visual environment, with a confidence level of $90\%$.

\begin{figure*}[htbp]
\centering
\subfloat[]
{
  \label{subfig8.a}
  \includegraphics[width=0.32\textwidth]{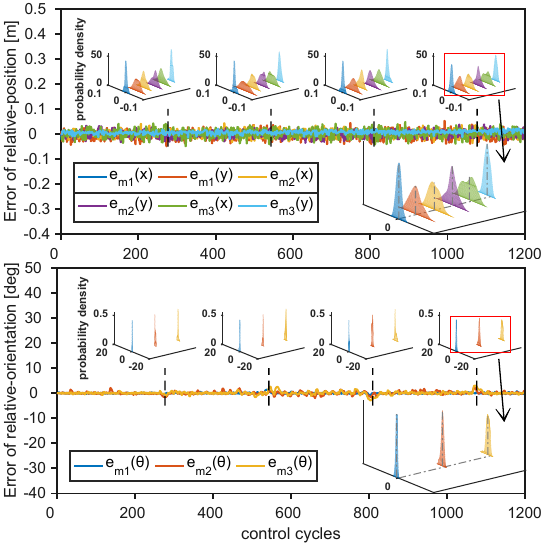}
}
\subfloat[]
{
  \label{subfig8.b}
  \includegraphics[width=0.32\textwidth]{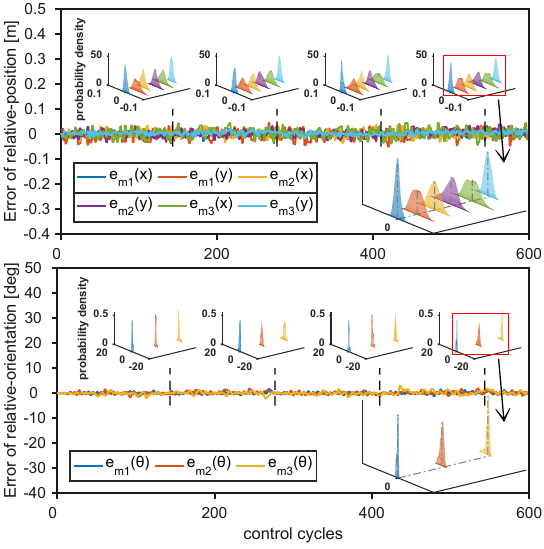}
}
\subfloat[]
{
  \label{subfig8.d}
  \includegraphics[width=0.32\textwidth]{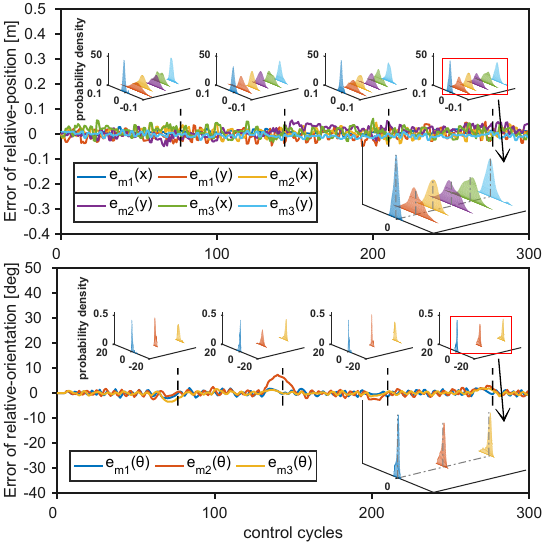}
}
\\
\subfloat[]
{
  \label{subfig8.e}
  \includegraphics[width=0.32\textwidth]{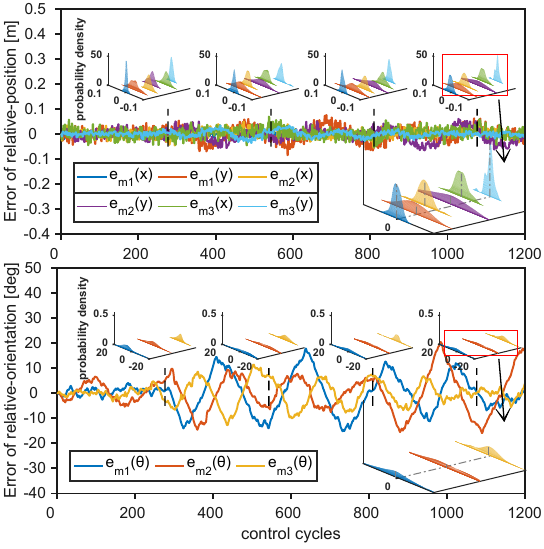}
}
\subfloat[]
{
  \label{subfig8.f}
  \includegraphics[width=0.32\textwidth]{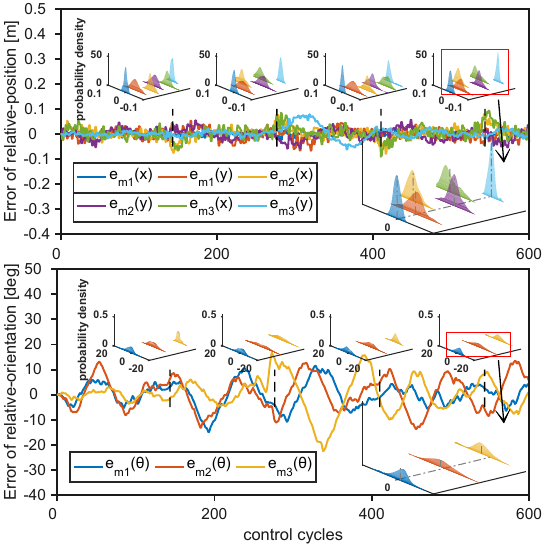}
}
\subfloat[]
{
  \label{subfig8.h}
  \includegraphics[width=0.32\textwidth]{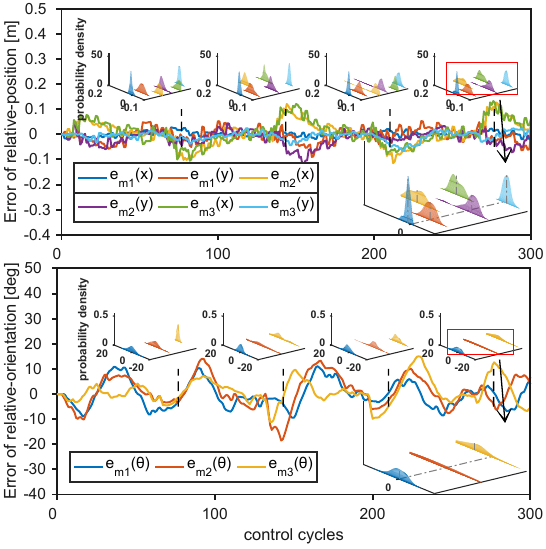}
}
\\
\subfloat[]
{
  \label{subfig8.i}
  \includegraphics[width=0.32\textwidth]{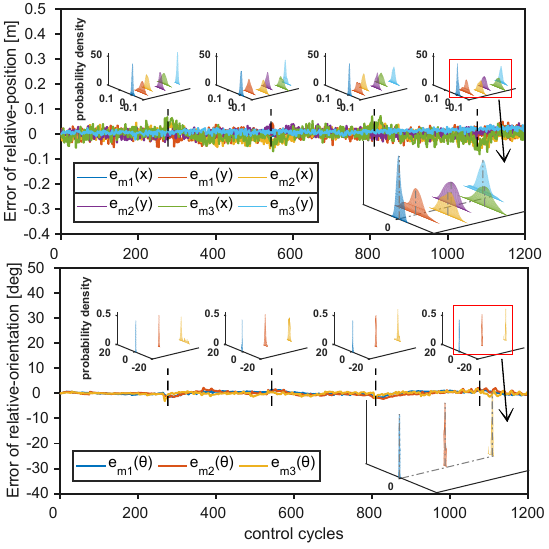}
}
\subfloat[]
{
  \label{subfig8.j}
  \includegraphics[width=0.32\textwidth]{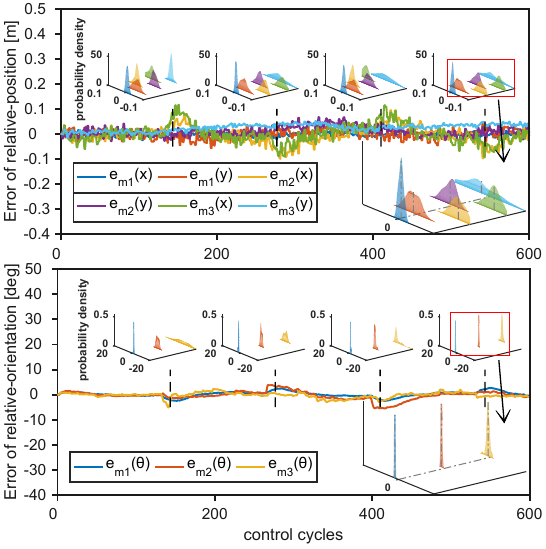}
}
\subfloat[]
{
  \label{subfig8.l}
  \includegraphics[width=0.32\textwidth]{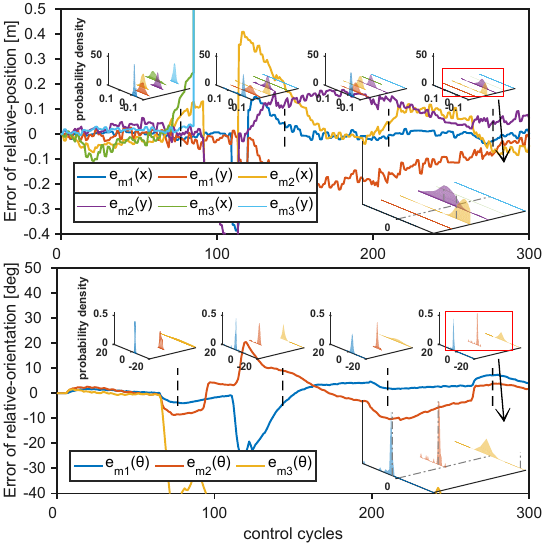}
}
\caption{Plots of Group 1 simulations: robots’ relative position and angular orientation errors of our approach ((a)-(c)), the discrete-time leader-follower method \cite{cruz2016leader} ((d)-(f)), and the continuous-time leader-follower method \cite{wang2023optical} ((g)-(i)), with $T=0.1s$ and $\bar{v}_{m,\hat{t}}=0.05m/s,0.1m/s$ and $0.2m/s$.}
\label{Fig_8}
\end{figure*}

\begin{figure*}[htbp]
\centering
\subfloat[]
{
  \label{subfig9.a}
  \includegraphics[width=0.32\textwidth]{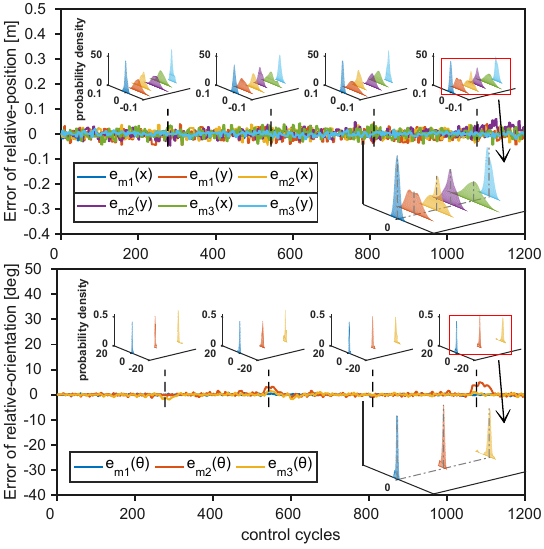}
}
\subfloat[]
{
  \label{subfig9.b}
  \includegraphics[width=0.32\textwidth]{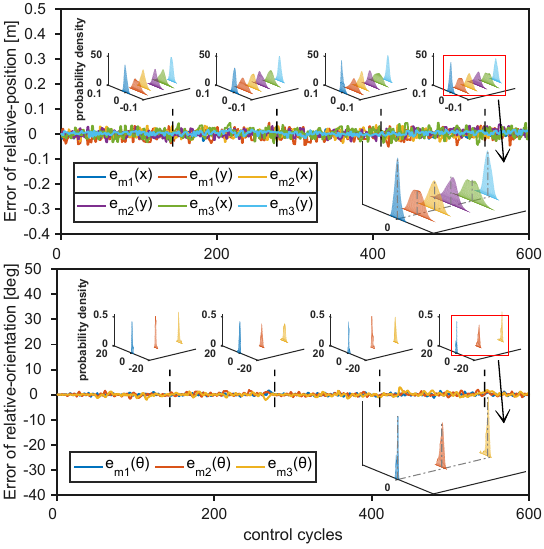}
}
\subfloat[]
{
  \label{subfig9.d}
  \includegraphics[width=0.32\textwidth]{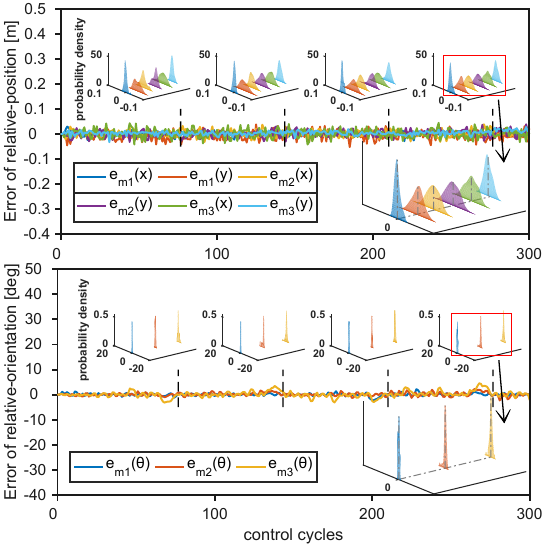}
}
\\
\subfloat[]
{
  \label{subfig9.e}
  \includegraphics[width=0.32\textwidth]{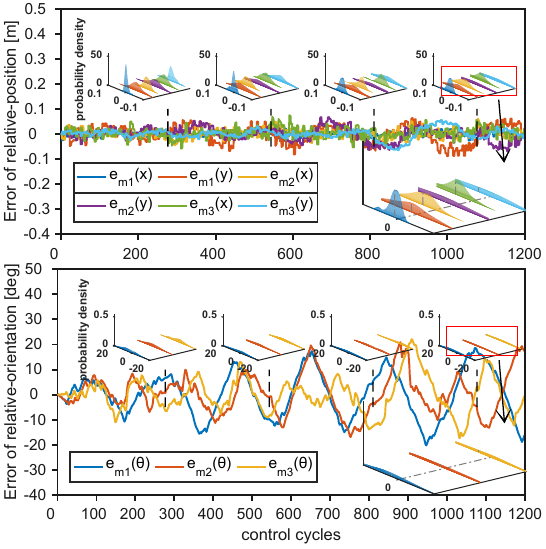}
}
\subfloat[]
{
  \label{subfig9.f}
  \includegraphics[width=0.32\textwidth]{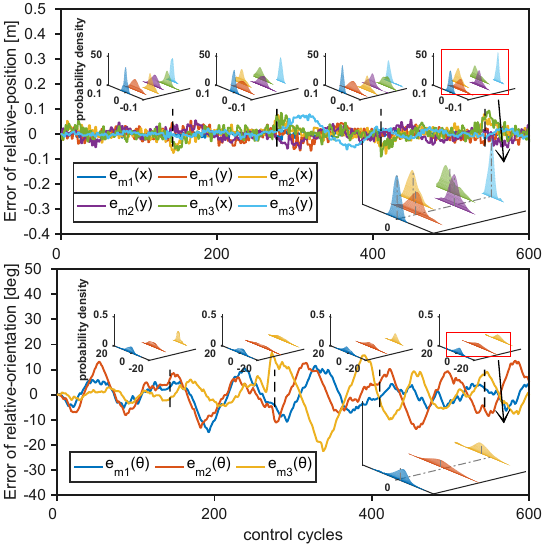}
}
\subfloat[]
{
  \label{subfig9.h}
  \includegraphics[width=0.32\textwidth]{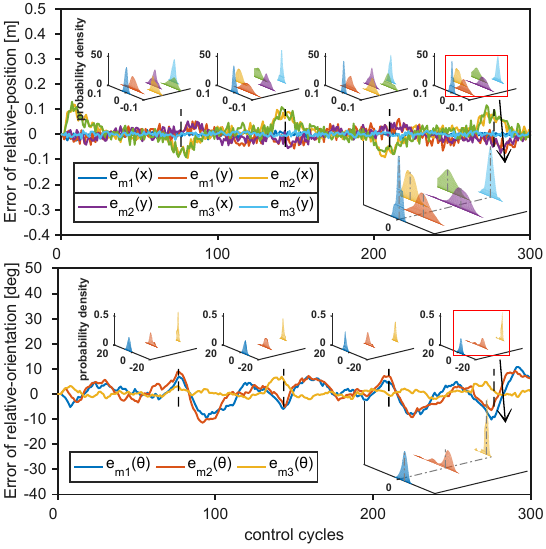}
}
\\
\subfloat[]
{
  \label{subfig9.i}
  \includegraphics[width=0.32\textwidth]{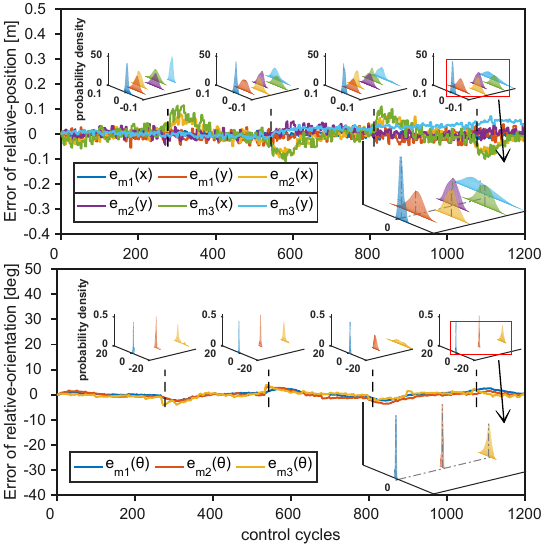}
}
\subfloat[]
{
  \label{subfig9.j}
  \includegraphics[width=0.32\textwidth]{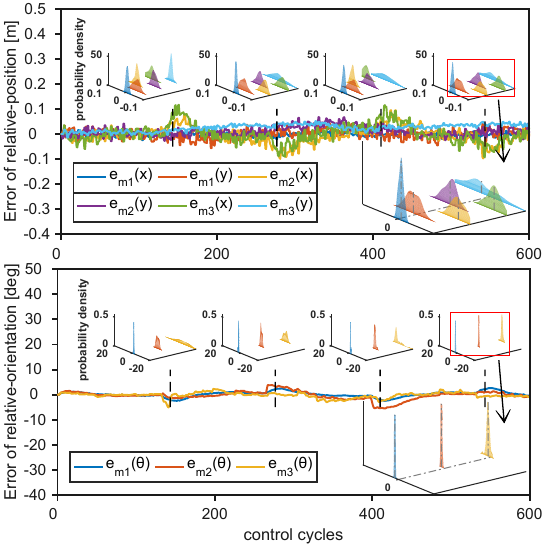}
}
\subfloat[]
{
  \label{subfig9.l}
  \includegraphics[width=0.32\textwidth]{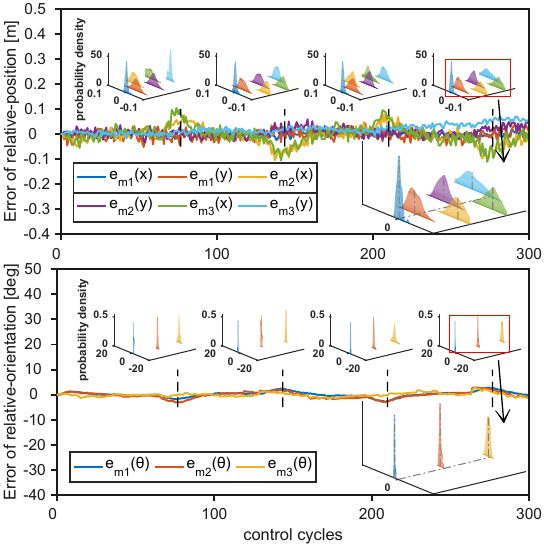}
}
\caption{Plots of Group 2 simulations: robots’ relative position and angular orientation errors of our approach ((a)-(c)), the discrete-time leader-follower method \cite{cruz2016leader} ((d)-(f)), and the continuous-time leader-follower method \cite{wang2023optical} ((g)-(i)), with $\bar{v}_{m,\hat{t}}=0.1m/s$ and $T=0.05s,0.1s$ and $0.2s$.}
\label{Fig_9}
\end{figure*}

For each combination of $\bar{\mathbf{\textbf{v}}}_{m,\hat{t}}=\left[\bar{v}_{m,\hat{t}},\bar{\omega}_{m,\hat{t}}\right]^T$ and $T$, 50 simulations were conducted to capture the statistical variations of formation errors. Figs. \ref{Fig_8} and \ref{Fig_9} show the formation errors for each combination of $\bar{\mathbf{\textbf{v}}}_{m,\hat{t}}$ and $T$ in Groups 1 and 2, respectively, for our method and the two benchmark methods. Each subfigure in Figs. \ref{Fig_8} and \ref{Fig_9} presents the traces of a typical formation error trajectory randomly selected from the 50 simulation runs. This trajectory was chosen without deliberate bias and reflects the typical behavior observed across all trials: other trajectories also exhibit similar error variation patterns. 

Each subfigure also includes the probability distribution of formation errors obtained from all 50 simulations at selected control cycles to illustrate the statistical variation of errors. To focus on the most challenging navigation maneuver within the S-shaped path, we selected the control cycles during which the preplanned target angular velocity $\bar{\omega}_{m,\hat{t}}$ underwent significant changes at the most curvy parts of the path, as marked by the vertical dashed lines in Figs. \ref{Fig_8} and \ref{Fig_9}. Kernel Density Estimation (KDE) \cite{chen2017tutorial} was employed to estimate the probability distributions of formation errors at these selected control cycles. Additionally, each subfigure also features an amplified zoom-in view of the error distribution in the last selected control cycle to highlight the increased susceptibility of the robot formation to deformation after extended movement. The dotted lines within these amplified views mark the mean formation error over the 50 simulations.

As shown by the traces in Figs. \ref{subfig8.a}-\ref{subfig8.d} and Figs. \ref{subfig9.a}-\ref{subfig9.d}, the typical formation error trajectories of our approach oscillate around the zero axis without exhibiting divergence throughout the navigation process. The amplified formation error distributions in these figures display a pronounced peak near the mean value across all combinations of $\bar{\mathbf{\textbf{v}}}_{m,\hat{t}}$ and $T$ in both groups. Notably, the mean values of the formation errors remain close to the zero axis, indicating that the formation errors neither experience substantial fluctuations over time nor increase significantly as $\bar{\mathbf{\textbf{v}}}_{m,\hat{t}}$ and $T$ vary.

From Figs. \ref{subfig8.a} to \ref{subfig8.d}, as the target linear velocity $\bar{v}_{m,\hat{t}}$ increases from $0.05m/s$ to $0.2m/s$, the maximum relative position error and angular orientation error of the formation error trajectories with our approach increase slightly from $\pm0.052m$ to $\pm0.061m$ and from $\pm 3.01 \degree$ to $\pm7.35\degree$, respectively. From Figs. \ref{subfig9.a} to \ref{subfig9.d}, as the control cycle duration $T$ increases from $0.05s$ to $0.2s$,  the formation error remains below $\pm0.053m$, while the maximum relative angular orientation error remains below $\pm4.54\degree$. These variations of $\bar{v}_{m,\hat{t}}$ and $T$ do not significantly degrade the formation maintenance performance of our approach. This is attributed to our approach's optimization of formation error correction at each control cycle, regardless of variations in $\bar{v}_{m,\hat{t}}$ and $T$, as both parameters are explicitly incorporated into the objective function. Additionally, the hold-and-hit framework synchronizes control cycles and execution instants of control inputs across all robots, ensuring that actual operations align with theoretical models without introducing additional errors.

In Figs. \ref{subfig8.e}-\ref{subfig8.h} and \ref{subfig9.e}-\ref{subfig9.h}, the formation error trajectories of the discrete-time leader-follower method \cite{cruz2016leader} exhibit significant fluctuations over time, particularly at the selected control cycles marked by vertical dashed lines. Although these trajectories do not ultimately diverge, the peak fluctuations may compromise the robot team’s ability to maintain a rigid formation.

Moreover, as shown in Figs. \ref{subfig8.e}-\ref{subfig8.h}, increasing values of $\bar{\mathbf{\textbf{v}}}_{m,\hat{t}}$ result in amplified formation error distributions exhibiting progressively greater skewness relative to zero. This is due to the approximations of the slave’s angular velocities in the control law design of \cite{cruz2016leader}. Conversely, in Figs. \ref{subfig9.e}-\ref{subfig9.h}, the formation errors do not exhibit significant fluctuations as a function of the control cycle duration $T$, which can be attributed to the incorporation of $T$ within the control law design of the discrete-time leader-follower method \cite{cruz2016leader}, ensuring compatibility with the ROS platform. The distributions of the relative angular orientation errors display a broader spread compared to our approach, indicating greater variability in relative angular orientation errors.

\begin{figure}[!t]
\centering
\includegraphics[width=2.5in]{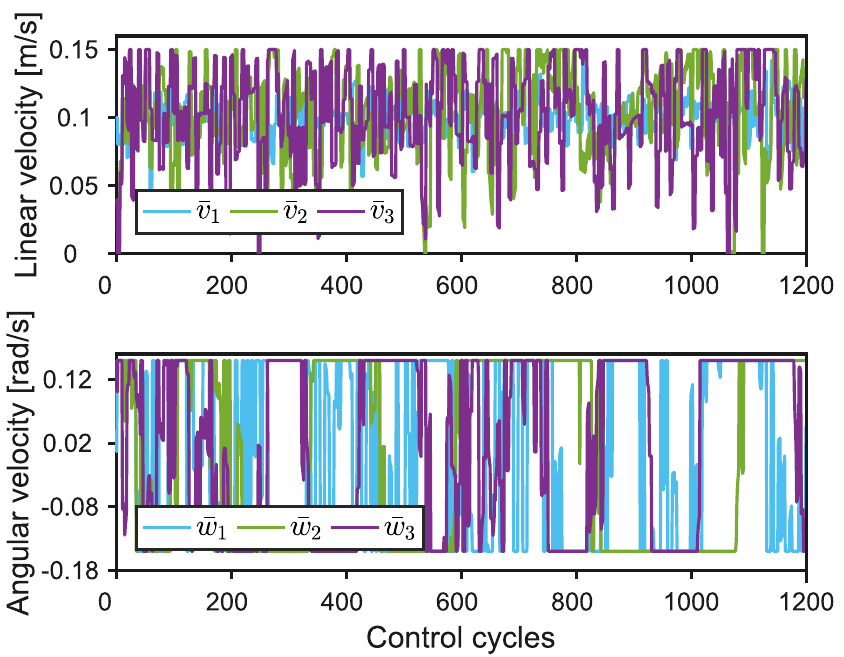}
\caption{Plots of a Group 2 simulation: the slaves’ target linear and angular velocities using the discrete-time leader-follower method \cite{cruz2016leader} with $\bar{v}_{m,\hat{t}}=0.1m/s$ and $T=0.05s$.}
\label{fig_10}
\end{figure}

As seen in Fig. \ref{subfig9.e}, when $\bar{v}_{m,\hat{t}}=0.1m/s$ and $T=0.05s$, the formation error using the discrete-time leader-follower method \cite{cruz2016leader} is exceedingly large. We further examine the target velocities $\bar{\mathbf{\textbf{v}}}_{i,\hat{t}}=\left[\bar{v}_{i,\hat{t}},\bar{\omega}_{i,\hat{t}}\right]^T$ for the slaves under these conditions. As shown in Fig. \ref{fig_10}, a considerable portion of the target linear velocity $\bar{v}_{i,\hat{t}}$ and angular velocity $\bar{\omega}_{i,\hat{t}}$ reach the maximum target velocity $\pm\bar{\mathbf{\textbf{v}}}_{max}$. This indicates that the slaves are often unable to fully execute the initially computed commands due to velocity limits, which hinders their ability to effectively correct formation errors. 

We next examine the continuous-time lead-follower method in \cite{wang2023optical}. As illustrated in Figs. \ref{subfig8.i}-\ref{subfig8.l} and \ref{subfig9.i}-\ref{subfig9.l}, when the target velocity $\bar{\mathbf{\textbf{v}}}_{m,\hat{t}}$ increases from $0.05m/s$ to $0.2m/s$ in Group 1 and the control cycle duration $T$ increases from $0.05s$ to $0.2s$ in Group 2, the formation performance deteriorates significantly. This deterioration is attributed to the continuous-time leader-follower method \cite{wang2023optical} not being fully compatible with the ROS platform. As either $\bar{\mathbf{\textbf{v}}}_{m,\hat{t}}$ or $T$ increases, the control law’s effectiveness rapidly declines. In Fig. \ref{subfig8.l}, the formation error trajectories diverge after fewer than 100 cycles.

Furthermore, as illustrated by the amplified light blue distributions in Figs. \ref{subfig8.i}-\ref{subfig8.l} and Figs. \ref{subfig9.i}-\ref{subfig9.l}, the formation errors of the slave positioned to the right of the master exhibit a more dispersed distribution compared to those of the other slaves. This suggests that the continuous-time leader-follower method \cite{wang2023optical} has limited ability to correct formation errors among followers alongside the master, further undermining its robustness in maintaining rigid formation.

Overall, the simulation results indicate that our approach is more effective in maintaining rigid formation while navigating curvilinear paths under error-free network conditions than the methods in \cite{cruz2016leader} and \cite{wang2023optical}. This effectiveness validates the accuracy of our theoretical analysis. 

\subsection{Robustness Evaluation}\label{subsec-IV(B)}

We next evaluate our approach under lossy communication. Recall that in our approach, $p$ is the probability that a slave receives control inputs from the master through the wireless network before the instant $\hat{t}+d$ in each control cycle. Due to the inherent uncertainty of wireless networks, $p$ may not always be accurately measured. We investigate both the cases where the ground-truth $p$ is known (i.e., measurable) and unknown. As will be demonstrated, our approach remains effective in rigid formation navigation regardless of whether the exact $p$ is available.

We conducted simulations using the same TurtleBot robots in the virtual environment as described in Section \ref{subsec-IV(A)}. The task remained rigid formation navigation along the S-shaped path as shown in Fig. \ref{fig_5a}. To simulate lossy communication conditions between the robots, we varied the success probability $p$ for command delivery. Since the robots are positioned close to each other, the probability of the control inputs failing to arrive before the instant $\hat{t}+d$ is generally low, regardless of whether TCP or UDP protocols are used for wireless communication (note: even if UDP is used, the lower-layer WiFi protocol may retransmit lost packets several times until they are successfully received; so even UDP may not exhibit very high $p$ after the retransmissions of MAC-layer ARQ). With that in mind, we varied $p$ from $0.9$ to $0.5$ to simulate lossy communication conditions. 

For all experiments, we fixed the $\bar{v}_{m,\hat{t}}$ to $0.1m/s$ and $T$ to $0.1s$. We varied the angular velocity $\bar{\omega}_{m,\hat{t}}$ within the range $\left[-0.1,0.1\right]rad/s$ to navigate the S-shaped path, following the pattern shown in the example in Fig. \ref{fig_6}. All experiments were divided into two groups to separately evaluate the effectiveness of our approach under conditions where the success probability $p$ is or is not known. 

\textbf{Group 1} examines how the formation error varies with different values of $p$ when the exact $p$ is known. In this group, the exact $p$ values ranging from $0.9$ to $0.5$ are substituted into (\ref{eq028}) to calculate the optimal target velocity $\bar{\mathbf{\textbf{v}}}_{i,\hat{t}}$ for the slaves. {\hfill $\square$}

\textbf{Group 2} examines whether our approach remains effective when $p$ is not known and therefore its value cannot be substituted into (\ref{eq028}). In this case, we set $p$ to 1 in (\ref{eq028}) regardless of the unknown ground-truth $p$. {\hfill $\square$}

For both groups, for each value of $p$, 50 simulations were conducted to capture the statistical variations of the formation errors. Other parameters in the DEM control law were kept consistent with those in Section \ref{subsec-IV(A)}, including the delay $d=0.5$, the weight matrix $\mathbf{\textbf{W}}=diag(1,1,1)$, and the noise power coefficient $\rho=1.4153\times10^{-5}$.

\begin{figure*}[htbp]
\centering
\subfloat[]
{
  \label{subfig11.a}
  \includegraphics[width=0.32\textwidth]{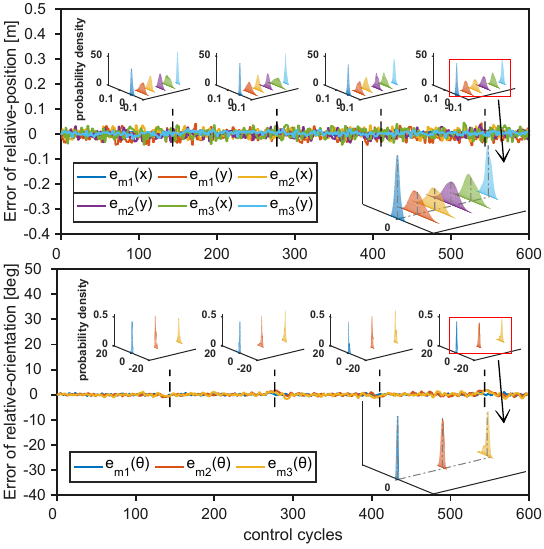}
}
\subfloat[]
{
  \label{subfig11.c}
  \includegraphics[width=0.32\textwidth]{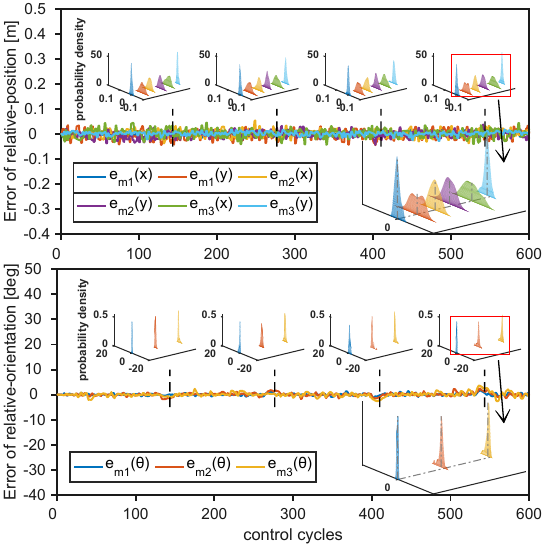}
}
\subfloat[]
{
  \label{subfig11.e}
  \includegraphics[width=0.32\textwidth]{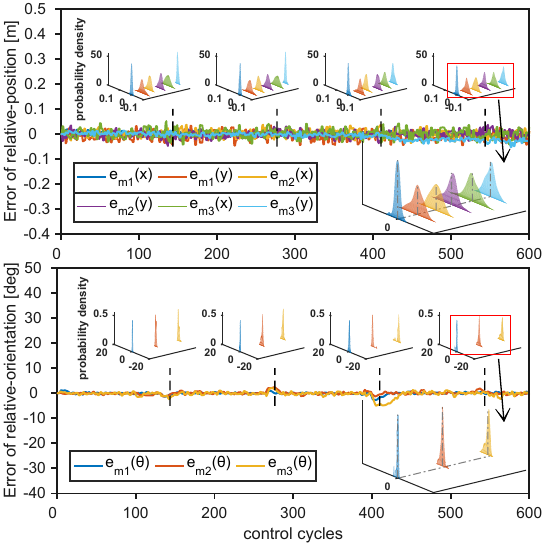}
}
\caption{Plots of Group 1 simulations: robots’ relative position and angular orientation errors of our approach, with $p=0.9,0.7$, and $0.5$ ((a)-(c)).}
\label{Fig_11}
\end{figure*}

\begin{figure*}[htbp]
\centering
\subfloat[]
{
  \label{subfig12.a}
  \includegraphics[width=0.315\textwidth]{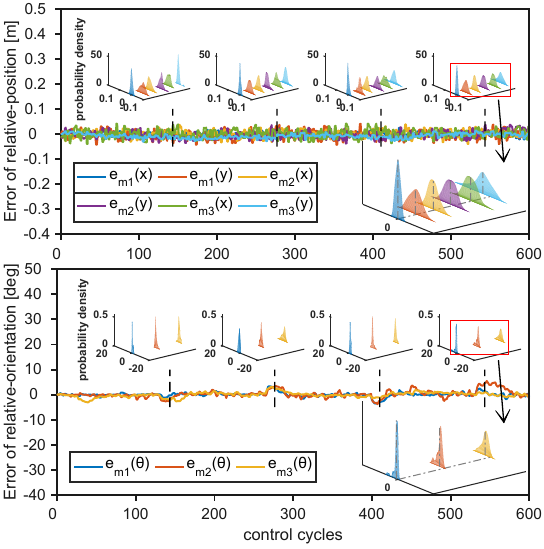}
}
\subfloat[]
{
  \label{subfig12.c}
  \includegraphics[width=0.315\textwidth]{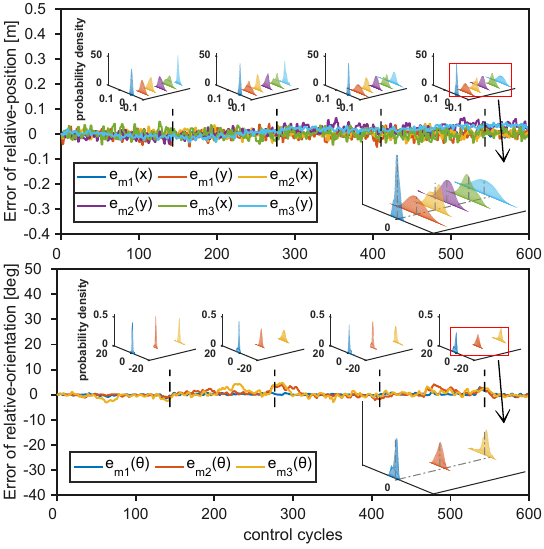}
}
\subfloat[]
{
  \label{subfig12.e}
  \includegraphics[width=0.315\textwidth]{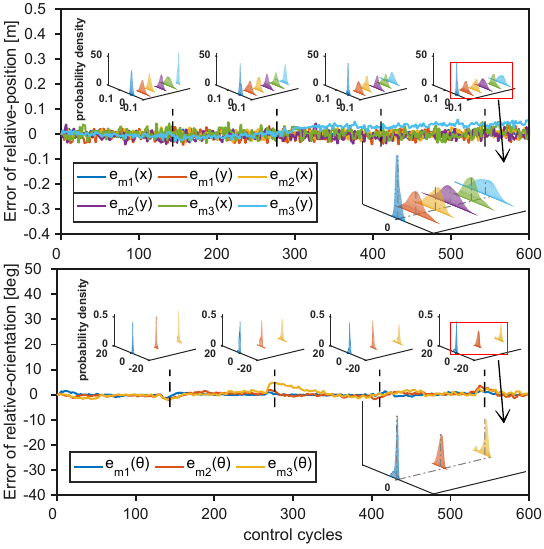}
}
\caption{Plots of Group 2 simulations: robots’ relative position and angular orientation errors of our approach, with $p=0.9,0.7$, and $0.5$ ((a)-(c)). The exact value of $p$ is assumed unknown to the robots.}
\label{Fig_12}
\end{figure*}

Figs. \ref{Fig_11} and \ref{Fig_12} show the formation errors for each value of $p$ in Groups 1 and 2. As with Figs. \ref{Fig_8} and \ref{Fig_9}, the dashed lines in these figures mark the control cycles when the S-shaped path curves sharply. As shown in Figs. \ref{subfig11.a}–\ref{subfig11.e} and \ref{subfig12.a}–\ref{subfig12.e}, when the success probability $p$ for command delivery decreases from $0.9$ to $0.5$, the formation errors remain close to the zero axis without divergence throughout the navigation in both groups.

In Group 1, as shown in Fig. \ref{subfig11.a}, when $p=0.9$, the relative positional error stays below $\pm0.054m$ and the relative orientation error stays below $\pm4.24\degree$. As $p$ decreases from $0.9$ to $0.5$, the formation performance deteriorates only slightly. When $p=0.5$, as shown in Fig. \ref{subfig11.e}, the maximum relative positional error remains below $\pm0.058m$, and the maximum relative angular orientation error stays under $\pm4.91\degree$. The amplified formation error distributions at the selected control cycles exhibit a pronounced peak near the mean value for all values of $p$. The mean values remain close to zero, indicating that the expected formation errors neither fluctuate drastically over time nor increase significantly as $p$ decreases. This robustness is attributed to our DEM control law, which optimizes the expected formation error at each control cycle by incorporating $p$ into the objective function. 

In Group 2, as shown in Figs. \ref{subfig12.a}–\ref{subfig12.e}, when $p$ decreases from $0.9$ to $0.5$, the maximum relative position error remains below $\pm0.064m$, and the maximum relative angular orientation error stays below $\pm8.72\degree$. Formation maintenance in Group 2 performs somewhat worse than in Group 1. This is because the DEM control law (\ref{eq028}) is always implemented with $p=1$, rather than the ground-truth $p$. The amplified formation error distributions also exhibit a slightly broader spread compared to those in Group 1, indicating increased spread in formation errors for Group 2. Despite this, the mean values remain close to zero, showing that the expected formation errors do not deteriorate significantly over time or as $p$ decreases. Although the unknown $p$ may lead to larger errors in each cycle, the errors can be observed at the beginning of the next cycle, and correction can be made then, preventing system divergence during navigation. 

Overall, the simulation results indicate that our approach remains effective under lossy communication conditions, regardless of whether the success probability $p$ for command delivery is precisely known. This further demonstrates the robustness of our approach.

\section{Experiments}\label{sec-V}
To further validate our approach, we applied it to real-world experiments. Hold-and-hit and DEM were implemented on physical TurtleBot robots communicating via a WiFi network. We assumed the success probability for command delivery to be $p=1$ in (\ref{eq028}), accounting for the fact that the ground-truth success probability for command delivery may not always be precisely measured and may be dynamically changing in Wi-Fi networks (see discussion in Section \ref{subsec-IV(B)}).

Recall that the DEM control law requires the noise power coefficient $\rho$, which varies across different robots and operating environments. This measurement process is time-consuming and could itself be error-prone. Notably, the measured $\rho$ in the simulation environment in Section \ref{sec-IV} is very small. Therefore, we set $\rho$ to 0 in (\ref{eq028}) without accurately measuring the ground-truth $\rho$ through extensive real-world experiments. This simplification makes our approach easier to apply in real-world scenarios, although it may potentially reduce performance. However, as will be demonstrated, our approach remains effective even if the ground-truth $\rho$ is unknown.

\begin{figure}
\centering
\subfloat[]
{
  \label{subfig13.0}
  \includegraphics[width=0.23\textwidth]{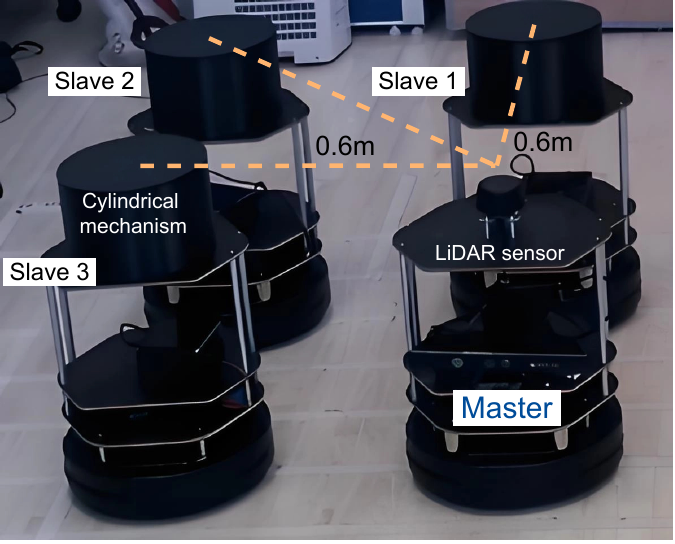}
}
\subfloat[]
{
  \label{subfig13.1}
  \includegraphics[width=0.23\textwidth]{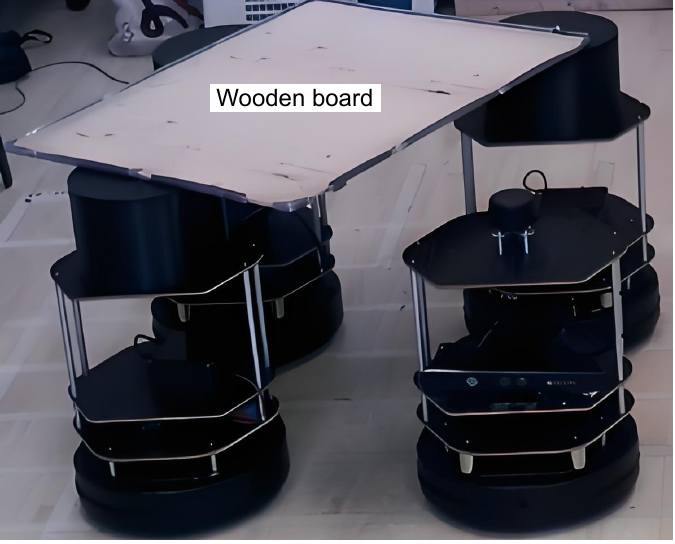}
}
\caption{A real-world MRS was developed to demonstrate the cooperative transportation scenario. (a) Four physical TurtleBots were arranged in a square formation. (b) The team of robots collectively lifted a wooden board.}
\label{fig_13}
\end{figure}

We deployed a team of four physical TurtleBots \cite{TurtleBot3} within a $2.4m \times 2.7m$ rectangular experimental area to cooperatively transport an object along an S-shaped path. As shown in Fig. \ref{subfig13.0}, the robots were initially arranged in a square formation with length of $0.6m$ on each side. The robots were tasked with transporting a wooden board measuring $0.9m \times 0.7m$ from an origin to a target location while preventing the wooden board from falling during navigation.

As shown in Fig. \ref{subfig13.0}, the robot layout and sensor setup were the same as those in the simulations in Section \ref{sec-IV}. Each robot was unaware of its absolute position within the experimental area. Each robot operated within a local ROS environment running on an Intel NUC 11, powered by the TurtleBot platform. A WiFi network was established among the robots, with the master configured in Access Point (AP) mode and the slaves in Station (STA) mode. To enable collective payload lifting, we installed a cylindrical mechanism atop each slave robot. A wooden board was manually placed on the cylindrical mechanisms of the slaves.

\begin{figure}
\centering
\subfloat[]
{
  \label{fig_14}
  \includegraphics[width=0.24\textwidth]{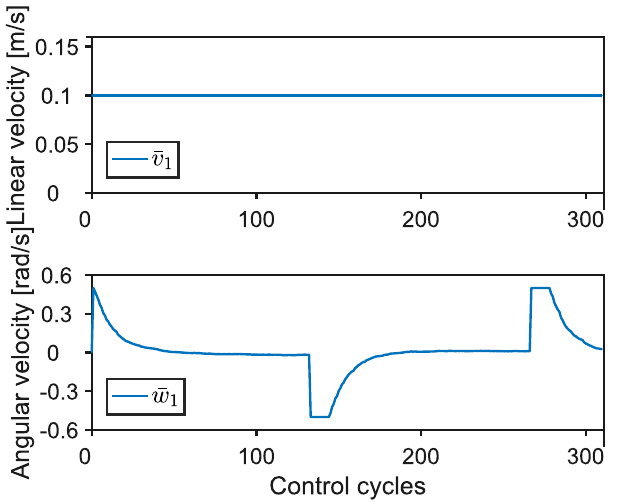}
}
\subfloat[]
{
  \label{fig_15}
  \includegraphics[width=0.23\textwidth]{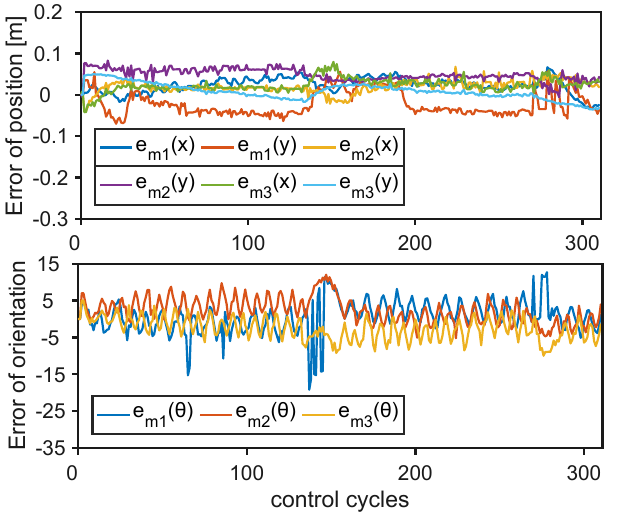}
}
\caption{(a) The setting of $\bar{v}_{m,\hat{t}}=0.1m/s$ and $\bar{\omega}_{m,\hat{t}} \in \left[-0.5,0.5\right]rad/s$ for the real-world cooperative transportation experiment. (b) Plots of the cooperative transportation experiment: robots’ relative position and angular orientation errors of our approach.}
\label{fig_14_update}
\end{figure}

We fixed the linear velocity $\bar{v}_{m,\hat{t}}=0.1m/s$ and varied the angular velocity $\bar{\omega}_{m,\hat{t}}$ within the range $\left[-0.5,0.5\right]rad/s$ to navigate the S-shaped path, as elaborated in Fig. \ref{fig_14}. The experiment lasted for 312 control cycles, each with a duration of $T=0.1s$. The maximum target velocity for the slaves was set to $\bar{\mathbf{\textbf{v}}}_{max}=1.5max\left|\bar{\mathbf{\textbf{v}}}_{m,\hat{t}}\right|$. Other parameters in the DEM control law remained the same as those in Section \ref{subsec-IV(A)}, including the delay $d=0.5$ and the weight matrix  $\mathbf{\textbf{W}}=diag(1,1,1)$.

Fig. \ref{fig_15} shows the formation errors for the real-world cooperative transportation scenario. The formation error trajectory oscillates around the zero axis without exhibiting divergence throughout the navigation process. The team of robots successfully transported the wooden board from the origin to the target position at $(0.06m,2.65m)$ following the S-shaped path without dropping the wooden board. A demo video of the experiment is available at: \url{https://youtu.be/GjbsF1HBkLw}. The maximum relative position error in the formation error trajectory stays below $\pm 0.069m$, while the maximum relative angular orientation error stays under $\pm 19.15\degree$. The formation maintenance performance in the real-world scenario is somewhat worse than in the simulations in Section IV. There are two possible reasons. First, the DEM control law (\ref{eq028}) was implemented with fixed parameters $\rho=0$ and $p=1$, ignoring the actual noise power coefficient $\rho$ and the success probability $p$. Second, and perhaps more importantly, the wooden board atop the robots may cause slight slipping or skidding in some units, introducing additional formation errors during both the "hold" and "hit" stages that are difficult to estimate.

Overall, the real-world experimental results indicate that although our approach does not achieve optimal performance when the noise power coefficient $\rho$ and success probability $p$ for command delivery are unknown, it remains effective in the cooperative transportation scenario under wireless communication conditions.

\section{Conclusion}\label{sec-VI}
This paper presents a discrete-time communication-control approach that addresses the challenge of maintaining rigid formation for untethered nonholonomic robots navigating curvilinear paths over wireless networks. The proposed method incorporates a “hold-and-hit” framework to synchronize robot movements and employs a Discrete-time Error Minimization (DEM) control law to reduce formation errors during motion. Fully compatible with the Robot Operating System (ROS) platform, the approach is designed to be robust against wireless network delays and packet loss. These features significantly enhance the practicality of untethered multi-robot rigid formation navigation in real-world applications, such as cooperative transportation in industrial environments. 

We provide a comprehensive investigation of the hold-and-hit framework and the DEM control law through theoretical analysis, simulations, and experiments. The theoretical study focuses on characterizing the formation error of nonholonomic robots when applying the hold-and-hit framework. Building on this foundation, we investigate the formation error correction within each control cycle during navigation using the DEM control law. Notably, the DEM control law is seamlessly integrated with the hold-and-hit framework and explicitly accounts for wireless network uncertainties in its optimization objective function. This intra-cycle optimization approach introduces a novel perspective for analyzing hybrid communication-control systems.

Our simulations demonstrate the effectiveness of the hold-and-hit framework combined with the DEM control law in maintaining the rigid formation of nonholonomic robots navigating curvilinear paths, outperforming two baseline methods. Additionally, both simulations and experiments validate the robustness of our approach under lossy communication conditions and confirm its feasibility in real-world cooperative transportation scenarios, underscoring its broad applicability and practical advantages. While the intra-cycle optimization approach may demand more computational resources than conventional methods, our results show that this overhead is well within the capabilities of modern microprocessors.

Looking ahead, the hold-and-hit framework and the optimized nonholonomic motion paradigm proposed in this work open new avenues for advancements in Multi-Robot Systems (MRS). By enabling robust and generalizable rigid formation navigation for nonholonomic differential mobile robots, our approach is widely applicable and holds the potential to transform multi-robot autonomous collaboration. It further supports stable rigid formation navigation for other nonholonomic robots at large scales, such as car-like robots and multirotor drones, while the robots communicate wirelessly in real-world scenarios.

\appendices
\section{Implementation of hold-and-hit mechanism and DEM control law}\label{sec-App1}
\begin{figure}[!t]
\centering
\includegraphics[width=3.3in]{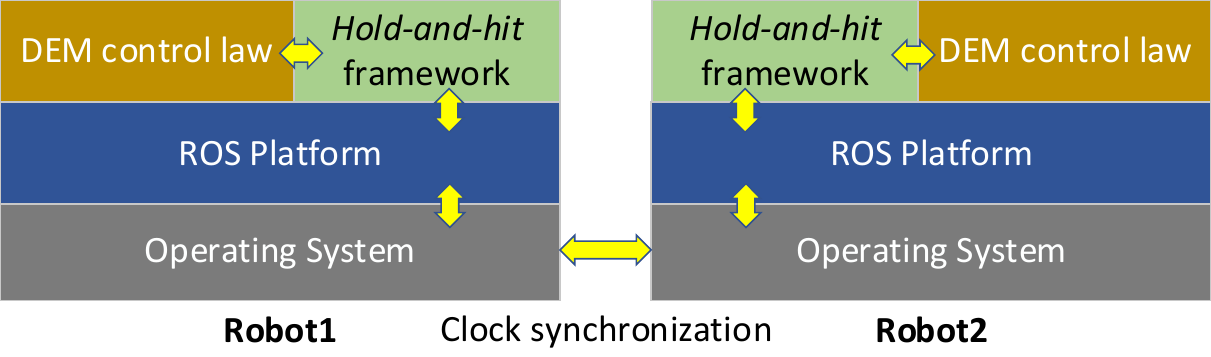}
\caption{The hold-and-hit framework operates as middleware situated between the ROS platform and the DEM control law.}
\label{fig_7}
\end{figure}

The proposed \textit{hold-and-hit} mechanism and DEM control law were implemented using ROS. Specifically, the hold-and-hit framework was developed as ROS-based middleware, while the DEM control law was realized as a ROS application. As illustrated in Fig. \ref{fig_7}, each robot operates within its own local ROS environment and communicates over a network. The hold-and-hit framework serves as general-purpose middleware between the ROS platform and the ROS application, enabling synchronous multi-robot motion even under lossy communication conditions. 

The framework employs the Precision Time Protocol (PTP) \cite{vallat2007clock} to synchronize the operating system clocks of all robots, achieving microsecond-level accuracy, which meets the synchronization requirements of the control cycle. All simulations in this paper were performed on a single physical PC, which inherently ensured clock synchronization across all robots. 

The ROS-based implementation of the hold-and-hit mechanism and the DEM control law from Section \ref{sec-IV} was compatible with the physical TurtleBots and seamlessly applied to real-world cooperative transportation. Since the WiFi NIC on the NUC 11 lacks the necessary hardware timestamp support for PTP, we set up an Ethernet LAN among the robots to synchronize their operating system clocks via PTP prior to the experiment. The achieved synchronization accuracy at the microsecond level meets the timing requirements of the control cycle.

The objective function of the DEM control law in (\ref{eq028}) was minimized using the L-BFGS algorithm \cite{nocedal1980updating}. This algorithm is particularly well-suited for deployment on low-cost edge hardware, such as the TurtleBot, due to its fast convergence and low memory consumption \cite{nocedal1980updating}. The optimal target velocity $\bar{\mathbf{\textbf{v}}}_{i,\hat{t}}$ is expected to be close to $\bar{\mathbf{\textbf{v}}}_{m,\hat{t}}$, as the slave robots must closely follow the master to maintain a rigid relative pose and the error at any time should remain small (hence the needed deviation of $\bar{\mathbf{\textbf{v}}}_{i,\hat{t}}$ with respect to $\bar{\mathbf{\textbf{v}}}_{m,\hat{t}}$ to correct for the error). As such, the optimization in (\ref{eq028}) was initialized at $\bar{\mathbf{\textbf{v}}}_{m,\hat{t}}$, which improves the convergence efficiency of the L-BFGS algorithm and helps avoid convergence to local optima. It is important to note that the optimization is performed once per control cycle. 

\bibliographystyle{IEEEtran}
\bibliography{Main}


\end{document}